\documentclass[sigconf]{acmart}

\usepackage{subfigure}
\usepackage{algorithmic}
\usepackage[linesnumbered,ruled]{algorithm2e}

\let\vec\mathbf
\newcommand{\floor}[1]{\lfloor #1 \rfloor}

\AtBeginDocument{%
  \providecommand\BibTeX{{%
    \normalfont B\kern-0.5em{\scshape i\kern-0.25em b}\kern-0.8em\TeX}}}

\copyrightyear{2021}
\acmYear{2021}
\setcopyright{acmlicensed}
\acmConference[GECCO '21]{2021 Genetic and Evolutionary Computation Conference}{July 10--14, 2021}{Lille, France}
\acmBooktitle{2021 Genetic and Evolutionary Computation Conference (GECCO '21), July 10--14, 2021, Lille, France}
\acmPrice{15.00}
\acmDOI{10.1145/3449639.3459383}
\acmISBN{978-1-4503-8350-9/21/07}

\begin{document}

\title{Self-Referential Quality Diversity\\Through Differential Map-Elites}

\author{Tae Jong Choi}
\affiliation{%
  \institution{Kyungil University}
  \city{Gyeongsan-si}
  \country{Republic of Korea}
}
\email{gry17@kiu.kr}

\author{Julian Togelius}
\authornote{The corresponding author}
\affiliation{%
  \institution{New York University}
  \city{Brooklyn, NY}
  \country{USA}
}
\email{julian@togelius.com}

\renewcommand{\shortauthors}{Tae Jong Choi and Julian Togelius}

\begin{abstract}
Differential MAP-Elites is a novel algorithm that combines the illumination capacity of CVT-MAP-Elites with the continuous-space optimization capacity of Differential Evolution. The algorithm is motivated by observations that illumination algorithms, and quality-diversity algorithms in general, offer qualitatively new capabilities and applications for evolutionary computation yet are in their original versions relatively unsophisticated optimizers. The basic Differential MAP-Elites algorithm, introduced for the first time here, is relatively simple in that it simply combines the operators from Differential Evolution with the map structure of CVT-MAP-Elites. Experiments based on 25 numerical optimization problems suggest that Differential MAP-Elites clearly outperforms CVT-MAP-Elites, finding better-quality and more diverse solutions.
\end{abstract}

\ccsdesc[500]{Computing methodologies~Continuous space search}
\ccsdesc[300]{Computing methodologies~Search methodologies}
\ccsdesc[100]{Computing methodologies~Artificial intelligence}

\keywords{Artificial intelligence, Evolutionary algorithms, Quality-diversity algorithms, Numerical optimization}

\maketitle

\section{Introduction}

Quality-diversity (QD) algorithms are a relatively recently devised class of evolutionary (or evolutionary-like) algorithms that search for a set of high-performing solutions that differs with a number of behavioral dimensions~\cite{pugh2016quality}. Such algorithms have a very large number of potential use cases that we are only starting to explore. However, existing QD algorithms are not, in general, very good optimizers, limiting their usefulness for some applications. In particular, real-valued optimization is a thoroughly studied area with an abundance of strong methods. If we could combine the behavior space exploration of QD algorithms with good real-valued optimization performance, we could extend the usefulness of QD algorithms to new domains.

This paper describes a new algorithm that combines the QD algorithm CVT-MAP-Elites \cite{vassiliades2017using} with the population-based optimization algorithm Differential Evolution (DE) \cite{storn1997differential,price2006differential}. CVT-MAP-Elites is a QD algorithm that keeps solutions in a centroidal Voronoi tesselation (CVT) where each cell contains the best-found solution in a particular part of behavior space. DE is a population-based optimization algorithm with good real-valued optimization performance. It is also remarkably simple in its original form.

\begin{figure}[t!]
 \centering
 \includegraphics[width=0.9\linewidth]{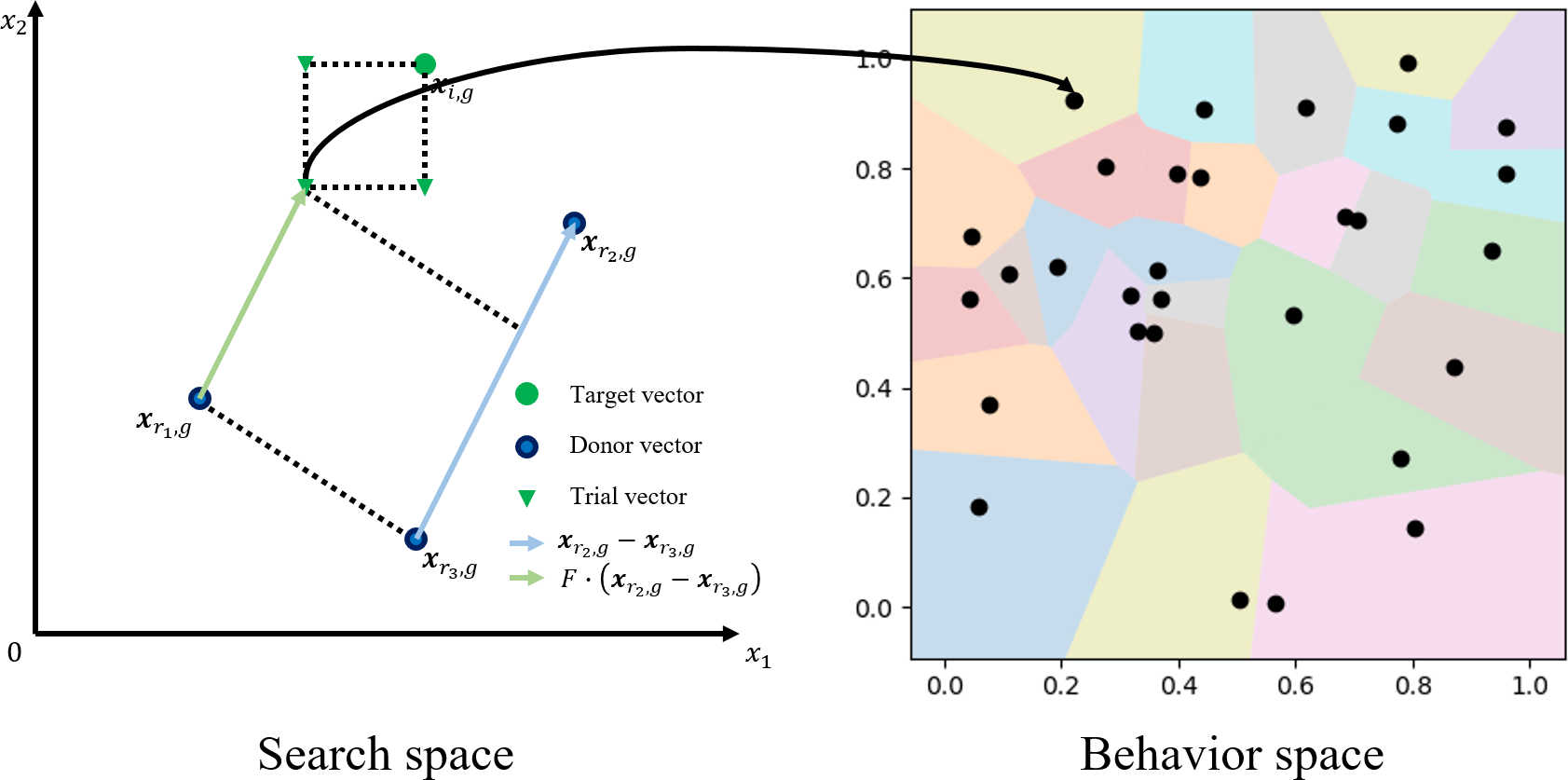}
 \caption[]{Search behavior of Differential MAP-Elites}
 \label{fig:search_behavior}
\end{figure}

The algorithm we propose here, Differential MAP-Elites, is a combination of the map structure of CVT-MAP-Elites with the operators from DE. CVT-MAP-Elites uses a mutation operator based on Gaussian distribution, which needs a carefully chosen step size. However, finding a suitable step size is a tedious and time-consuming process. DE uses a mutation operator based on linear combinations of vectors, which can automatically adjust its step size through randomly selected candidate solutions. Therefore, the proposed algorithm can achieve improved performance by adapting the operators from DE. An example of the search behavior of Differential MAP-Elites is shown in Fig. \ref{fig:search_behavior}.

We carried out empirical studies on 25 different and difficult numerical optimization problems \cite{suganthan2005problem} to evaluate the performance of Differential MAP-Elites. The experimental results suggest that Differential MAP-Elites clearly outperforms CVT-MAP-Elites, which finds better-quality and more diverse solutions.

\section{Background}

\subsection{Quality-diversity algorithms}

QD algorithms are algorithms that find a set of solutions that covers a space that is generally referred to as a behavior space~\cite{pugh2016quality}\footnote{Quality-diversity algorithms were developed in the context of artificial life and robot navigation, which explains the behavior terminology.}. In other words, these algorithms seek to find many different solutions (``behaviors'') that perform well. Two of the most well-known QD algorithms are Novelty Search with Local Competition~\cite{lehman2011evolving} and MAP-Elites~\cite{mouret2015illuminating}. However, there are many QD algorithms, including those that have taken the basic QD concept in new directions, such as Surprise Search~\cite{gravina2016surprise}, DeLeNoX~\cite{liapis2013transforming}, and Innovation Engines~\cite{nguyen2015innovation}.

\subsection{MAP-Elites and CVT-MAP-Elites}

MAP-Elites is a QD algorithm that has seen use in domains as different as robot locomotion planning~\cite{cully2015robots,nordmoen2019evolved}, game level generation~\cite{khalifa2018talakat,khalifa2019intentional,alvarez2019empowering,gravina2019procedural}, game strategy characterization~\cite{fontaine2019mapping}, and workforce scheduling~\cite{urquhart2018optimisation}. This algorithm requires that each individual can be evaluated not only for fitness but for at least one behavior characteristic (BC) as well. These BCs are dimensions along which you want to achieve differentiation. For example, when evolving a game level, the BCs could be related to the geometry of the level or whether it can be completed by certain agents~\cite{gravina2019procedural}. When evolving a robot gait, the BCs could be related to whether a particular leg was used in the gait~\cite{cully2015robots}. The output of the MAP-Elites algorithm is a set of high-performing (high fitness) individuals, with each individual being the most fit within a particular part of a space described by the BCs.

The fundamental data structure of MAP-Elites is a map that is a grid where individuals are placed depending on their BCs. Each cell contains the fittest individual within a particular range of BC values. Whenever a new individual is generated, it replaces the elite in that cell if it has a higher fitness. CVT-MAP-Elites~\cite{vassiliades2017using} instead relies on a CVT, which has the advantage that the granularity of a grid does not need to be determined (because there is no grid).

MAP-Elites and CVT-MAP-Elites, like several other QD algorithms, rely on simple and straightforward mutation and/or crossover for reproduction. While this is effective, there are many more sophisticated operators.

\subsection{CMA-MAP-Elites}

In a related (and mostly concurrent) development to the method presented in this paper, a different algorithm has been created to fuse MAP-Elites with Covariance Matrix Adaptation Evolution Strategies (CMA-ES)~\cite{fontaine2020covariance}. The resulting method, CMA-ME, employs multiple different ``emitters'' that perform CMA-ES-powered searches so as to improve the coverage of behavior space and the fitness of the best individuals. A full comparison between the two methods has not been performed yet, but we expect them to have their various strengths and weaknesses depending on the problem type. For comparison, algorithms derived from DE and from CMA-ES both place highly on continuous optimization benchmarks, but which one performs better depends largely on the nature of the particular benchmarks. Another related development is the directional variation operator, which relies on correlations between different elites to accelerate MAP-Elites~\cite{vassiliades2018discovering}.

\subsection{Differential Evolution}

DE~\cite{storn1997differential,price2006differential} is a population-based optimization algorithm that performs very well on many benchmarks and applications where the chromosome is expressed as a vector of real numbers. Variations and extensions of this algorithm frequently appear at or near the top of various continuous optimization competitions~\cite{das2016recent}.

DE searches for solutions by maintaining a population of $NP$ candidate solutions, and each candidate solution is a vector of $D$ real numbers. DE's optimization process can be divided into two stages: initialization and evolution. In the initialization stage, DE randomly distributes the population over the search space of a given problem. In the evolution stage, DE repeats generating $NP$ new candidate solutions and forming the next generation population by comparing the old and new candidate solutions. The evolution stage is terminated if a termination criterion is met.

\subsubsection{Initialization}

Let $\vec{x}_{i,g} = (x_{i,g}^{1}, \cdots, x_{i,g}^{D})$ denote the $i$th candidate solution at the generation $g$. Also, let $\vec{x}_{min}$ = ($x_{min}^{1}$, $\cdots$, $x_{min}^{D}$) and $\vec{x}_{max}$ = ($x_{max}^{1}$, $\cdots$, $x_{max}^{D}$) denote the maximum and minimum bounds, respectively. DE initializes each candidate solution according to

\begin{equation}
x_{i,0}^{j} \leftarrow x_{min}^{j} + rand_{i,j} \cdot (x_{max}^{j} - x_{min}^{j})
\end{equation}

\noindent
where $rand_{i,j}$ denotes the uniformly distributed random real number on the interval $[0,1]$.

\subsubsection{Mutation}

DE generates a set of $NP$ mutant vectors in the mutation phase. Each mutant vector denoted by $\vec{v}_{i,g}$ is generated by a linear combination of three randomly selected mutually distinct candidate solutions, $\vec{x}_{r_{1},g}$, $\vec{x}_{r_{2},g}$, and $\vec{x}_{r_{3},g}$, which can be formulated according to

\begin{equation}
\vec{v}_{i,g} \leftarrow \vec{x}_{r_{1},g} + F \cdot (\vec{x}_{r_{2},g} - \vec{x}_{r_{3},g})
\label{de_op1}
\end{equation}

\noindent
where $F$ denotes the scaling factor. This mutation operator explained above is called DE/rand/1, and there are many other variations and extensions in the literature of DE.

\subsubsection{Crossover}

DE generates a set of $NP$ trial vectors (new candidate solutions) in the crossover phase. Each trial vector denoted by $\vec{u}_{i,g}$ is generated by a combination of a target vector $\vec{x}_{i,g}$ and a mutant vector $\vec{v}_{i,g}$, which can be formulated according to

\begin{equation}
u_{i,g}^{j} \leftarrow \left\{ \begin{array}{ll}
v_{i,g}^{j} & \textrm{if $rand_{i,j} \leq CR$ or $j = j_{rand}$} \\
x_{i,g}^{j} & \textrm{otherwise}
\end{array} \right.
\label{de_op2}
\end{equation}

\noindent
where $CR$ denotes the crossover rate and $j_{rand}$ denotes the uniformly distributed random integer number on the interval $[1,D]$. This crossover operator explained above is called binomial crossover, which is the most frequently used in the literature of DE.

\subsubsection{Selection}

DE forms the next generation population by comparing all of the pairs of a target vector $\vec{x}_{i,g}$ and its corresponding trial vector $\vec{u}_{i,g}$. If a trial vector has a lower objective value (a higher fitness) than its corresponding target vector does, the trial vector is selected as a candidate solution for the next generation population, and the target vector is discarded; otherwise, vice versa. This can be formulated according to

\begin{equation}
\vec{x}_{i,g+1} \leftarrow \left\{ \begin{array}{ll}
\vec{u}_{i,g} & \textrm{if $f(\vec{u}_{i,g}) \leq f(\vec{x}_{i,g})$} \\
\vec{x}_{i,g} & \textrm{otherwise.}
\end{array} \right.
\end{equation}

\noindent
where $f(\vec{x})$ denotes the given problem to be minimized.


\section{Differential MAP-Elites}

Gaussian distribution mutations are a key operator for MAP-Elites and CVT-MAP-Elites to search for solutions. However, there are some limitations to the mutations as follows.

\begin{itemize}
  \item Gaussian distribution mutations have the sigma parameter that determines the step size. The sigma parameter is strongly dependent on the characteristics of a given problem, and thus, it needs to be carefully chosen. However, finding a suitable sigma parameter is a tedious and time-consuming process. Moreover, different sigma parameters may be preferred during different optimization processes.
  \item It is theoretically and empirically verified that mutations based on short-tailed distributions, e.g. Gaussian distribution, have relatively slow convergence in solving multimodal optimization problems \cite{yao1999evolutionary}. Since short-tailed distributions approach zero fast, the mutations have a low probability of generating long jumps. Therefore, MAP-Elites and CVT-MAP-Elites may have a poor ability to escape from a local optimum.
\end{itemize}

The core contribution of this paper, Differential MAP-Elites, is an algorithm that combines CVT-MAP-Elites with DE. This algorithm is designed to overcome the limitations of MAP-Elites and CVT-MAP-Elites outlined above and thereby to improve performance in a complex environment.

The basic operation of the canonical DE algorithm (DE/rand/1/bin) is remarkably simple. To create an offspring, three donor chromosomes are selected. The first donor is subtracted from the second, yielding a difference vector; this difference vector is then scaled and subtracted from a third chromosome. The resulting chromosome is then combined with the parent chromosome (target vector), which creates the offspring (trial vector). If the offspring is more fit than the parent chromosome, it replaces that chromosome. The magnitude and direction of the difference vector are strongly dependent on the location and dispersion of the population. If the donor chromosomes are located close to each other, the search step is narrow because the magnitude of the difference vector is small, whereas the search step is wide if the donor chromosomes are located far away from each other. Therefore, DE can automatically adjust its step size through randomly selected chromosomes, called a self-reference.

Differential MAP-Elites is a version of MAP-Elites. The core idea of this algorithm is simple: the operators are taken from DE, everything else from CVT-MAP-Elites. The pseudocodes of the algorithm are presented in Algs. \ref{alg:DME}, \ref{alg:CVT} and \ref{alg:AddToArchive}.

The proposed algorithm has the following inputs: an objective function $f(\vec{x})$, a behavior function $b(\vec{x})$, the dimension of a behavior space $N$, and the number of centroids $k$. The proposed algorithm divides the $N$-dimensional behavior space into $k$ homogeneous cells with a CVT \cite{ju2002probabilistic}. After that, the proposed algorithm creates an empty archive of size $k$. Accordingly, the proposed algorithm is ready to start the optimization process.

The optimization process can be divided into two stages: initialization and reproduction.
In the initialization stage, the proposed algorithm randomly distributes $G$ chromosomes over the search space of a given problem. After that, the performance ($f(\vec{x})$) and behavior ($b(\vec{x})$) of each chromosome are computed. With the computed behavior, the index of a centroid that is closest to the behavior of each chromosome is computed as well. If multiple chromosomes fit into the same cell (the same centroid index), the fittest one per cell remains. In the reproduction stage, the proposed algorithm repeats the following steps until a termination criterion is met:

\begin{itemize}
  \item Select four distinct non-empty cells from the behavior space.
  \item Obtain four chromosomes from the selected cells and use them as a target chromosome $\vec{x}_{i}$ and three donor chromosomes $\vec{x}_{r_1}$, $\vec{x}_{r_2}$, and $\vec{x}_{r_3}$.
  \item Obtain a trial chromosome $\vec{u}_{i}$ with the Eqns. \ref{de_op1} and \ref{de_op2}.
  \item Compute the performance and behavior of the trial chromosome and attempt to replace the elite in that cell if the trial chromosome has a higher fitness.
\end{itemize}

Many different termination conditions could be used, such as whether function evaluations exceed the maximum function evaluations, whether simulation time exceeds the maximum simulation time, or whether a certain number of cells are filled.


\begin{algorithm}[t!]
    \SetKwInOut{Input}{Input}
    \SetKwInOut{Output}{Output}

    \Input{Objective function $f(\vec{x})$, behavior function $b(\vec{x})$, the dimension of a behavior space $N$, the number of centroids $k$, scaling factor $F$, and crossover rate $CR$}
    \Output{Archive ($X, P$)}

    \tcp{Run CVT and get the centroids $C$}
    $C \leftarrow$ CVT\_Approximation($k$)\;
    $X \leftarrow$ create\_empty\_archive(k)\;
    \While{$i = 1 \rightarrow G$}{
        $\vec{x} \leftarrow$ create\_random\_solution()\;
        Add\_To\_Archive($\vec{x}$, $X$, $P$)\;
    }
    \While{$i = 1 \rightarrow I$}{
        \tcp{Select a target vector $\vec{x}$ and three donor vectors $\vec{r_{1}}, \vec{r_{2}}, \vec{r_{3}}$ ($\vec{x} \ne \vec{r_{1}} \ne \vec{r_{2}} \ne \vec{r_{3}}$)}
        $\vec{x}, \vec{r_{1}}, \vec{r_{2}}, \vec{r_{3}} \leftarrow$ selection(X)\;
        \tcp{Run DE/rand/1 mutation (Eq. \ref{de_op1})}
        $\vec{v} \leftarrow$ DE/rand/1($\vec{x}$, $\vec{r_{1}}$, $\vec{r_{2}}$, $\vec{r_{3}}$)\;
        \tcp{Run binomial crossover (Eq. \ref{de_op2})}
        $\vec{u} \leftarrow$ bin($\vec{x}$, $\vec{v}$)\;
        Add\_To\_Archive($\vec{u}$, $X$, $P$)\;
    }
    \Return $(X, P)$\;
    \caption{Differential MAP-Elites}
    \label{alg:DME}
\end{algorithm}

\begin{algorithm}[t!]
    \SetKwInOut{Input}{Input}
    \SetKwInOut{Output}{Output}

    \Input{The number of centroids $k$}
    \Output{Centroids $C$}
    
    \tcp{Generate $k$ random centroids}
    $C \leftarrow$ get\_sample\_Point($k$)\;
    \tcp{Generate $K$ random samples}
    $S \leftarrow$ get\_sample\_Point($K$)\;
    \While{$i = 1 \rightarrow iter_{max}$}{
        $I_{S} \leftarrow$ get\_closest\_centroid\_indices($C$, $S$)\;
        $C \leftarrow$ update\_centroids($I_{S}$)
    }
    \Return $C$\;
    \caption{CVT\_Approximation (Taken from \cite{ju2002probabilistic})}
    \label{alg:CVT}
\end{algorithm}

\begin{algorithm}[t!]
    \SetKwInOut{Input}{Input}

    \Input{Chromosome $\vec{x}$, solutions $X$, and performance $P$}
    
     \tcp{Compute the performance and behavior}
    $p \leftarrow f(\vec{x})$\;
    $b \leftarrow b(\vec{x})$\;
    \tcp{Compute the index of a centroid that is closest to the behavior}
    $I_{b} \leftarrow$ get\_closest\_centroid\_index($C$, $b$)\;
    \If{$P(I_{b}) =$ null or $P(I_{b}) < p$}
    {
        $X(I_{b}) \leftarrow \vec{x}$\;
        $P(I_{b}) \leftarrow p$\;
    }
    \caption{Add\_To\_Archive}
    \label{alg:AddToArchive}
\end{algorithm}


\section{Experiment settings}

\subsection{System configuration}

We used a PC with an AMD Ryzen Threadripper 2990WX and Microsoft Windows 10 Pro to carry out all the empirical studies in this paper. We used the Python programming language to implement all the test algorithms in this paper.

\subsection{Experiment domain}

\begin{figure}[t!]
 \centering
 \includegraphics[width=0.9\linewidth]{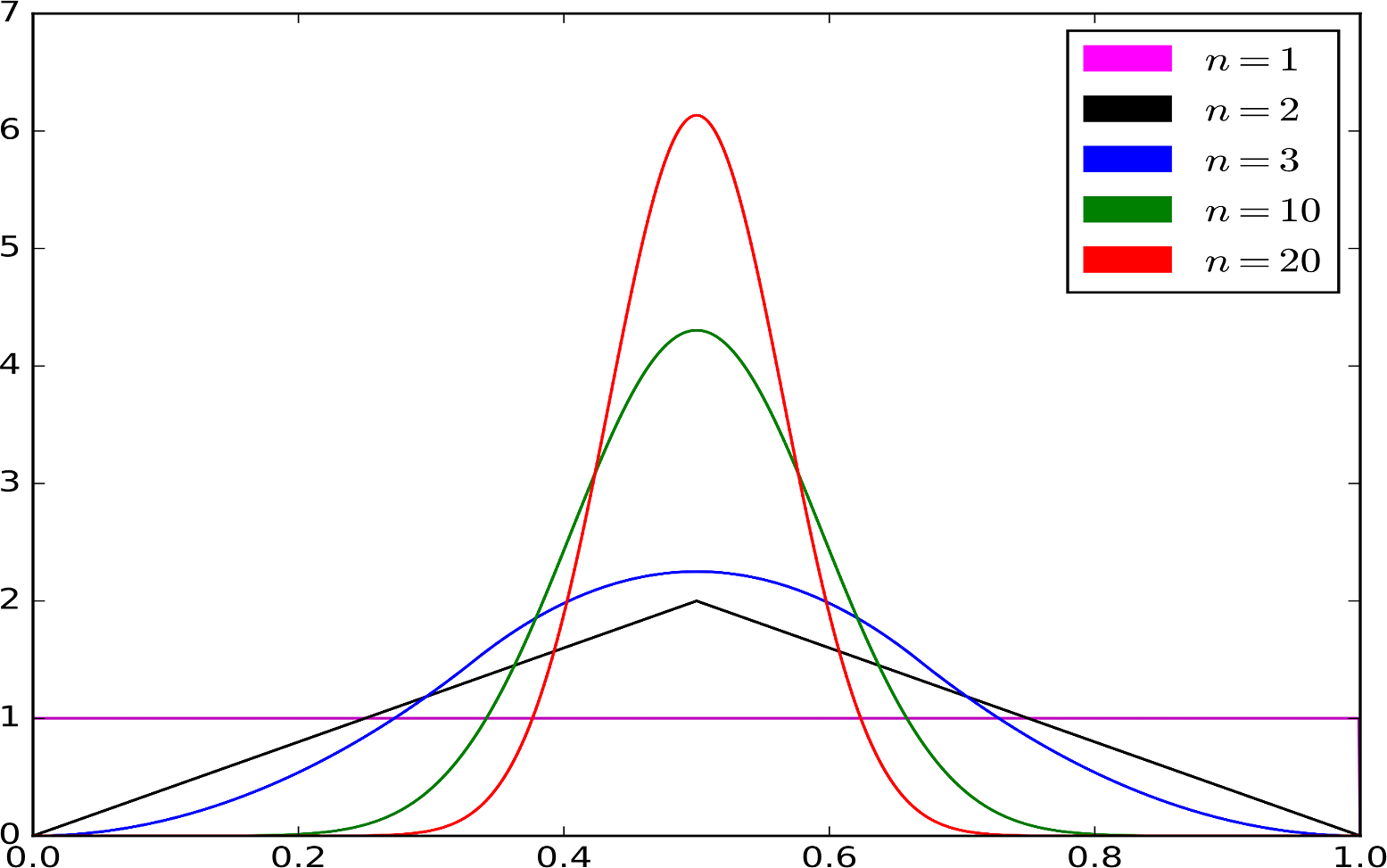}
 \caption[]{The effects of a linear projection are the same as a Bates distribution. As the dimension of a search space increases, the distribution of a distorted behavior space narrows.}
 \label{fig:Bates}
\end{figure}

To evaluate the performance of Differential MAP-Elites in a complex environment, we employed the experiment domain described in \cite{fontaine2020covariance}. The authors introduced a linear projection that maps from a high-dimensional search space to a low-dimensional behavior space, creating a highly distorted behavior space. In the behavior space, each BC is determined by all parameters in the search space, and each parameter equally contributes to its corresponding BC. The distribution of a distorted behavior space is the same as a Bates distribution if the mapping from a search space to a behavior space is the normalized sum of all parameters (Fig. \ref{fig:Bates}). As the dimension of a search space increases, the distribution of a distorted behavior space narrows, making it difficult to navigate all parameters to reach peaks. The authors showed that the standard MAP-Elites algorithm suffers from significant performance degradation in a highly distorted behavior space.

As for the objective function, we employed 25 different and difficult numerical problems taken from the CEC 2005 test suite \cite{suganthan2005problem}. The numerical problems have various properties, such as unimodal/multimodal, separable/non-separable, shifted, rotated, and/or fitness noise. The experiment settings for the numerical problems, such as the lower and upper bounds of search space, global optimum, and termination criterion, are the same as in \cite{suganthan2005problem}. Note that there are more recent test suites, such as the CEC 2013, 2014, and 2017 test suites. Still, we chose the CEC 2005 test suite because, unlike the others, it supports various search bounds which are useful in measuring the behavior space exploration capability of test algorithms.

As for the behavior function, we employed a linear projection from ${\rm I\!R}^{n}$ to ${\rm I\!R}^{2}$, computing the sum of the first half of elements from ${\rm I\!R}^n$ and the sum of the second half of elements from ${\rm I\!R}^n$. This can be formulated according to

\begin{equation}
b(\vec{x}) = \Big( \sum_{i=1}^{\floor{n/2}} clip \big( x^{i} \big), \sum_{i=\floor{n/2}+1}^{n} clip \big( x^{i} \big) \Big)
\end{equation}

\begin{equation}
clip \big( x^{i} \big) = \left\{ \begin{array}{ll}
x_{min}^{i}/x^{i} & \textrm{if $x^{i} \leq x_{min}^{i}$} \\
x^{i} & \textrm{if $x_{min}^{i} \leq x^{i} \leq x_{max}^{i}$} \\
x_{max}^{i}/x^{i} & \textrm{if $x_{max}^{i} \leq x^{i}$}
\end{array} \right.
\end{equation}

\noindent
The clip function keeps the contribution of each element $x^{i}$ to its corresponding BC within the bounds $[x_{min}^{i},x_{max}^{i}]$.

\subsection{Test algorithms}

We carried out empirical studies to evaluate the performance of Differential MAP-Elites and compared it with CVT-MAP-Elites \cite{vassiliades2017using}. The control parameters of the CVT-MAP-Elites algorithm were initialized according to the respective publication for a fair comparison, i.e., $k = 25,000$, $G = n \cdot 100$, and $\sigma^{i} = (x_{max}^{i} - x_{min}^{i})/300$. The control parameters of Differential MAP-Elites were initialized as the same as CVT-MAP-Elites, and $F = 0.5$ and $CR = 0.9$.

\subsection{Performance metrics}

\subsubsection{Function Error Value}

We used the function error value (FEV) to measure the accuracy of a test algorithm, which can be formulated according to

\begin{equation}
\textrm{FEV} = f(\vec{x}_{\text{best}}) - f(\vec{x}^{\star})
\end{equation}

\noindent
where $\vec{x}_{\text{best}}$ is the best-found solution and $\vec{x}^{\star}$ is the global optimum. A test algorithm has a higher accuracy as the FEV decreases.

\subsubsection{Coverage}

We used the ratio of non-empty cells to measure the behavior space exploration capability of a test algorithm. A test algorithm has a higher behavior space exploration capability as the ratio of non-empty cells increases.

\subsubsection{Statistical test}

We used the Wilcoxon rank-sum test with $\alpha = 0.05$ significance level \cite{demvsar2006statistical} to statistically compare the accuracy and behavior space exploration capability of two algorithms for each problem. The symbols used in this paper have the following meanings.

\begin{enumerate}
\item +: The CVT-MAP-Elites algorithm obtains a significantly better performance as compared to the proposed algorithm. \label{item:1}
\item =: The performance difference between the proposed algorithm and the CVT-MAP-Elites algorithm is not statistically significant. \label{item:2}
\item -: The proposed algorithm obtains a significantly better performance as compared to the CVT-MAP-Elites algorithm. \label{item:3}
\end{enumerate}

\section{Experiment results}

In this section, we present the experiment results. We labeled the proposed algorithm and CVT-MAP-Elites as CVT-DME and CVT-ME, respectively. Note that we measured the average best FEVs only for the two test functions $F_{7}$ and $F_{25}$ because these functions have the infinite bounds of search space.

Table \ref{tab:result1} presents the average best FEVs and coverages of 30 observations for each test function at 2 and 10 dimensions. As can be seen from the table, CVT-DME outperforms CVT-ME to obtain accurate and diverse solutions.

In the experiments at the 2 dimension, CVT-DME finds seven significantly more accurate solutions than CVT-ME, while CVT-ME finds two significantly more accurate solutions than CVT-DME. We can see a more interesting performance difference between the two algorithms from the coverage results where CVT-DME fills more behavior space than CVT-ME does on 22 test functions, statistically significant. Regarding the coverage results, CVT-DME clearly outperforms CVT-ME on the test functions from $F_{1}$ to $F_{6}$, $F_{8}$, and $F_{14}$. It is interesting to note that these functions have a broader search space ($\vec{x} \in [-100,100]^{n}$) than the other functions do. CVT-DME performs well, and its behavior space exploration capability is consistent on all the test functions ($\approx$ 92\%). Therefore, these results emphasize CVT-ME's weakness that its operators are ineffective if the search space of a given problem is broad.

In the experiments at the 10 dimension, CVT-DME finds seven significantly more accurate solutions than CVT-ME, while CVT-ME finds nine significantly more accurate solutions than CVT-DME. Although CVT-ME performs slightly better than CVT-DME in terms of the average best FEVs, CVT-DME fills more behavior space than CVT-ME does on 13 test functions, statistically significant.

\begin{figure*}[pt]
 \centering
 \subfigure[$F_{2}$ ($n = 10$)]{
  \includegraphics[scale=0.23]{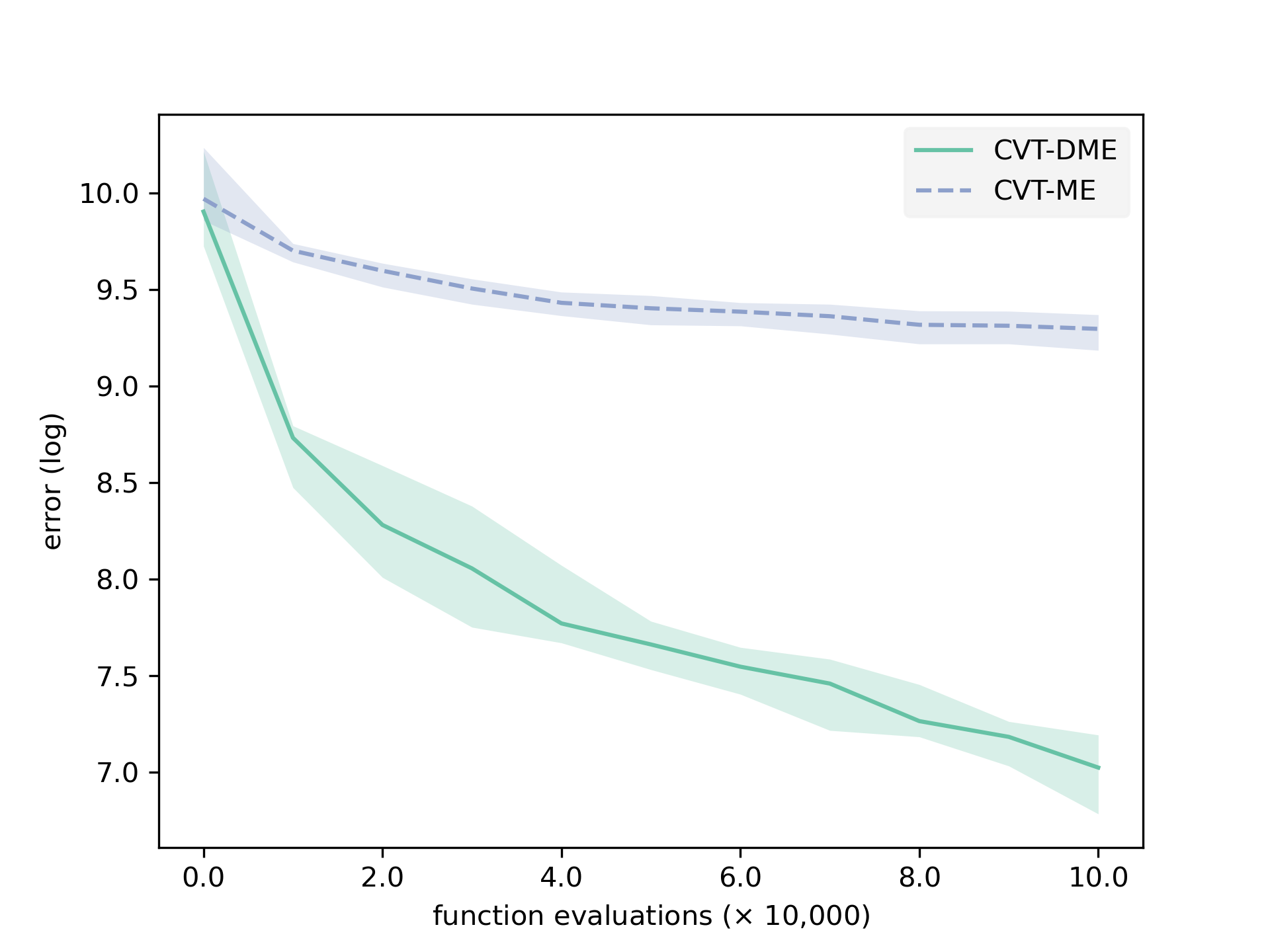}
   \label{fig:2_10_2}
   }
 \subfigure[$F_{4}$ ($n = 10$)]{
  \includegraphics[scale=0.23]{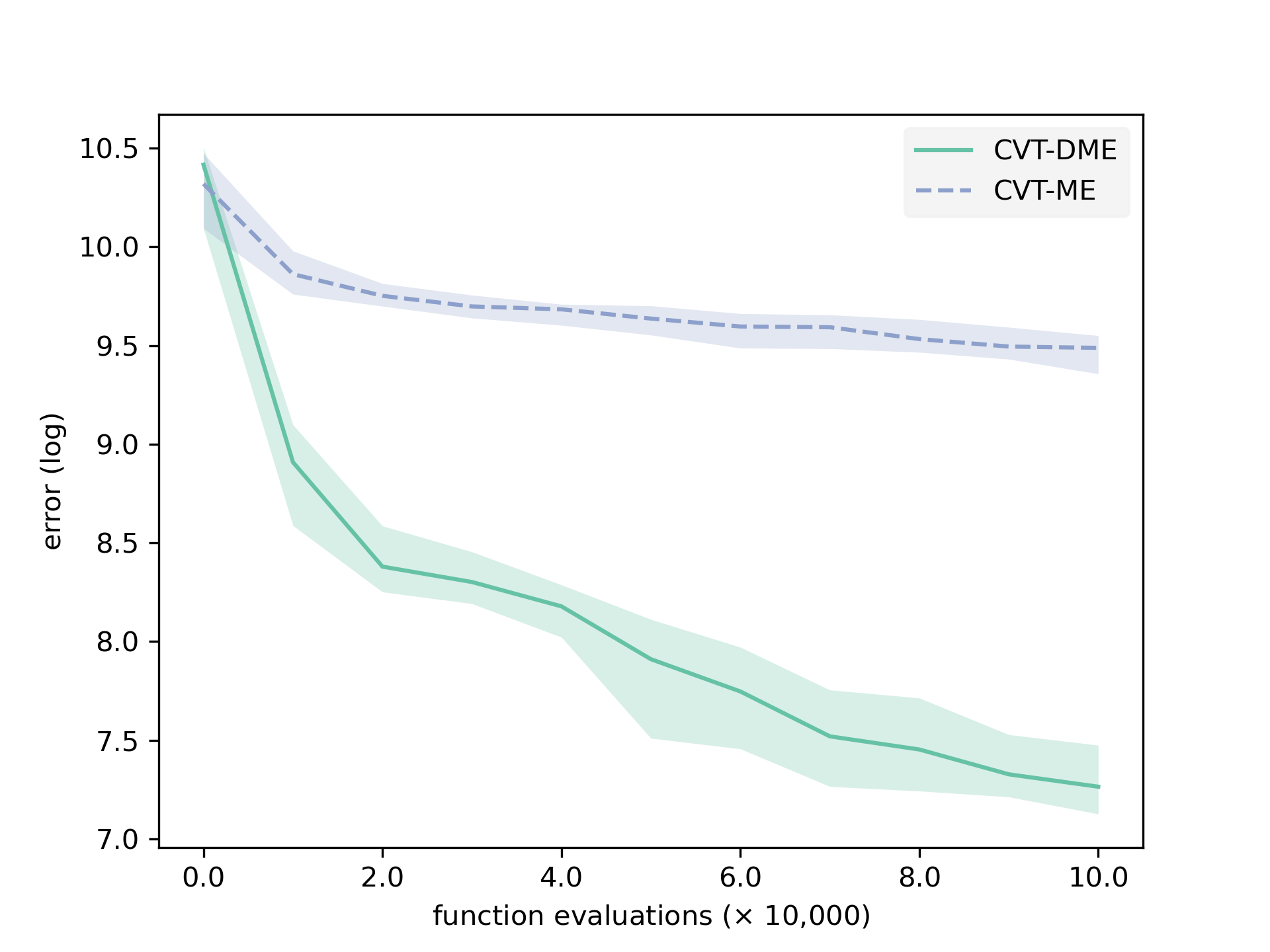}
   \label{fig:2_10_4}
   }
 \subfigure[$F_{6}$ ($n = 10$)]{
  \includegraphics[scale=0.23]{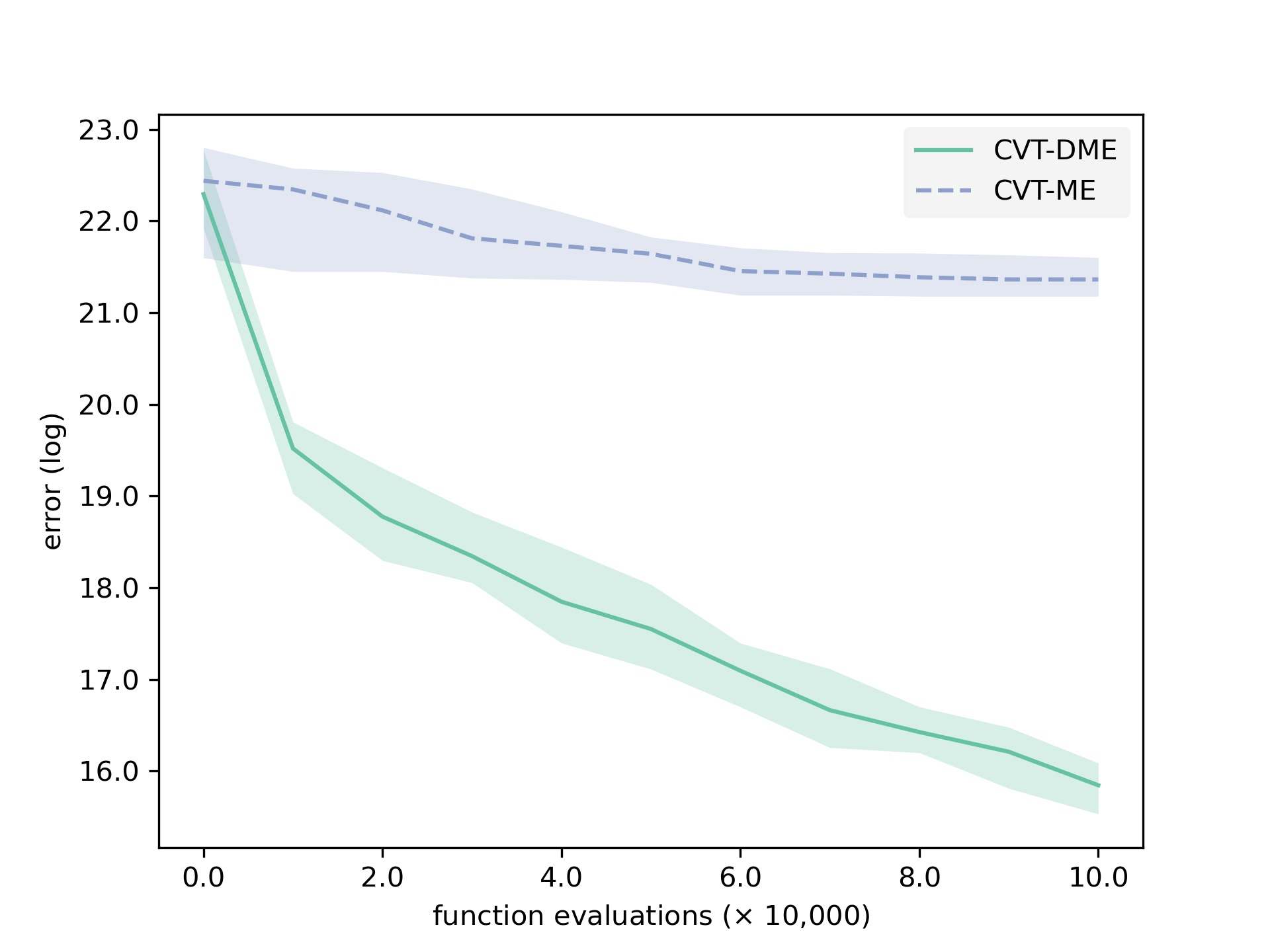}
   \label{fig:2_10_6}
   }
 \subfigure[$F_{8}$ ($n = 10$)]{
  \includegraphics[scale=0.23]{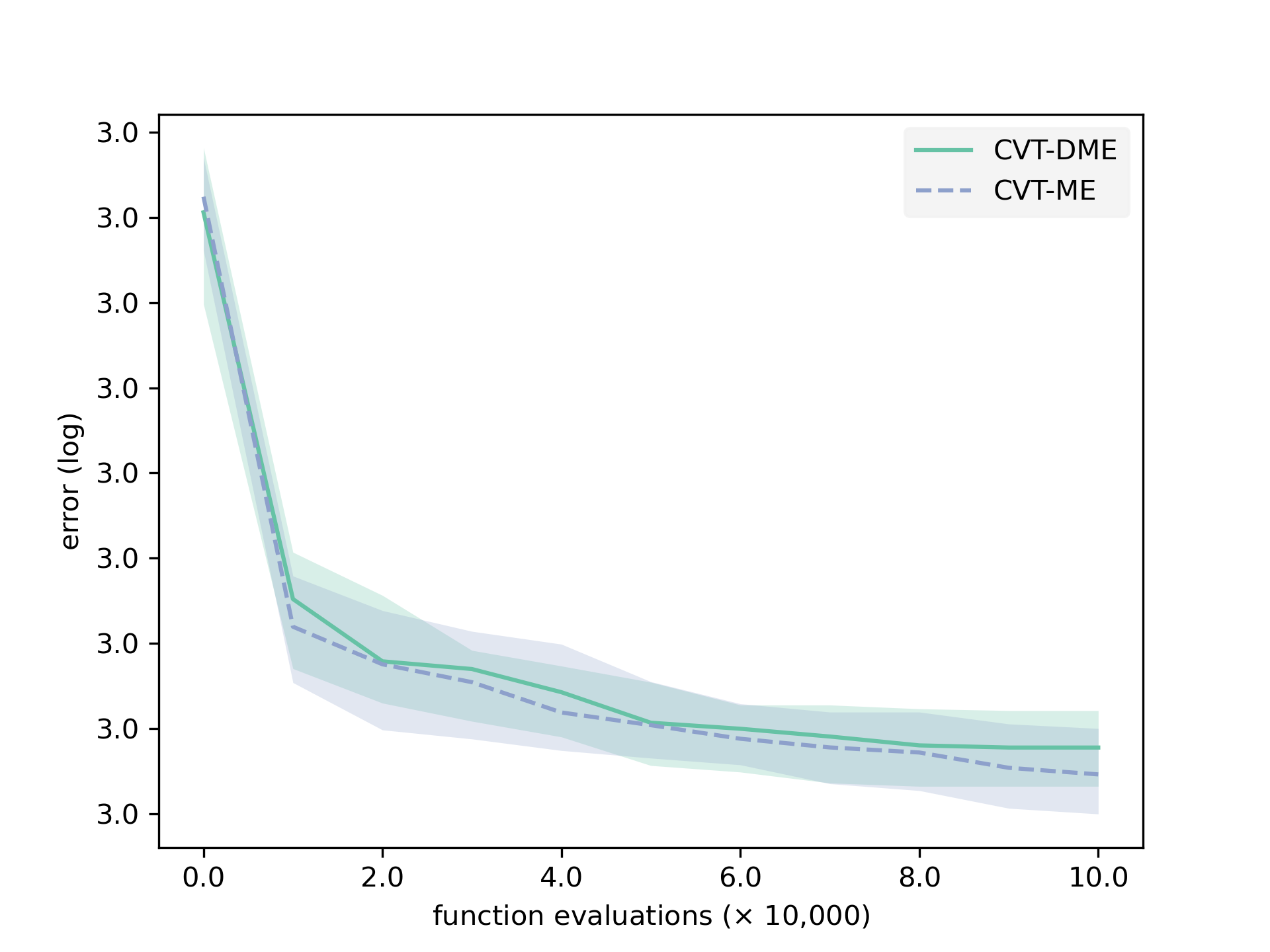}
   \label{fig:2_10_8}
   }
\subfigure[$F_{14}$ ($n = 10$)]{
  \includegraphics[scale=0.23]{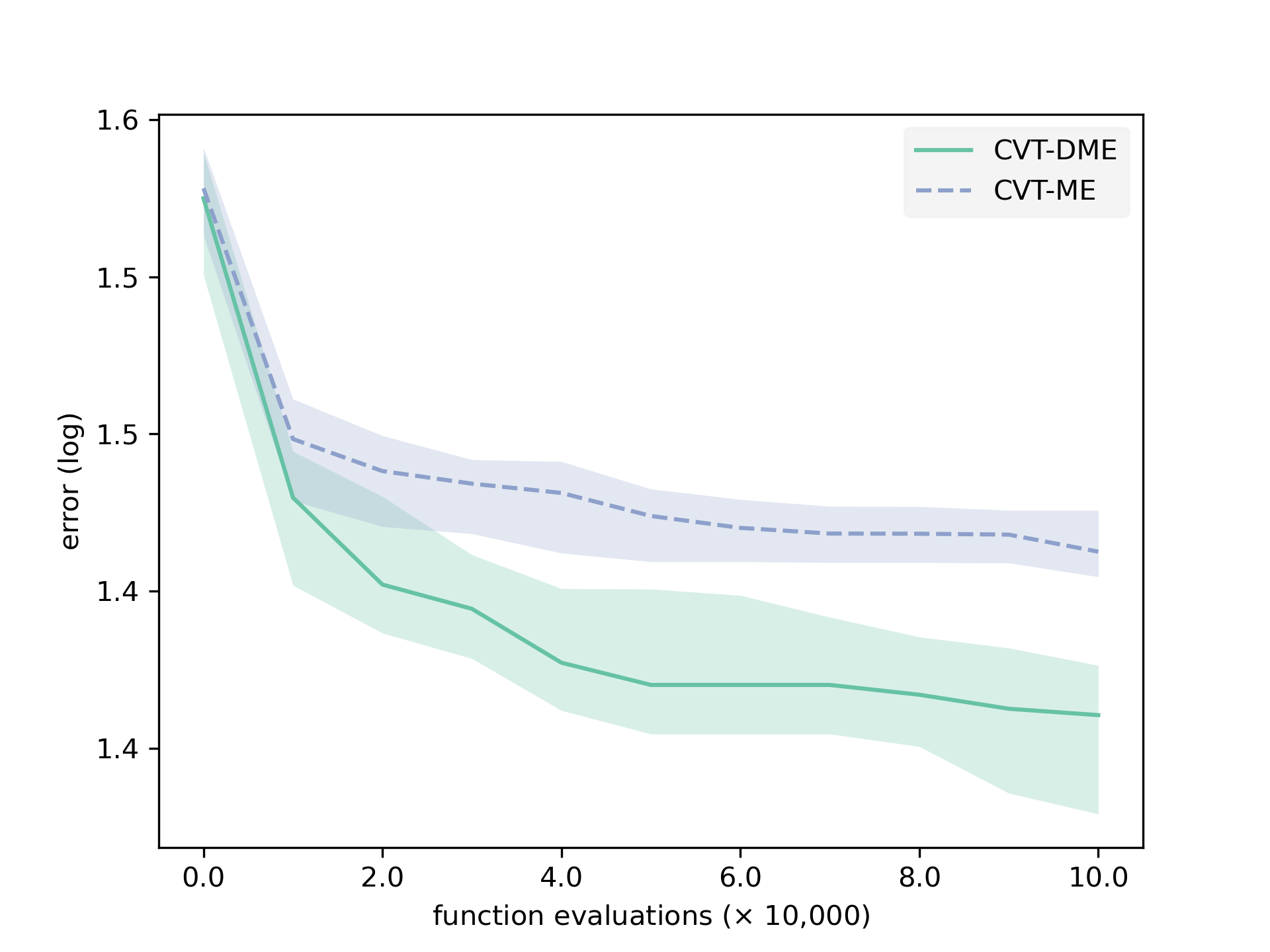}
   \label{fig:2_10_14}
   }
 \subfigure[$F_{18}$ ($n = 10$)]{
  \includegraphics[scale=0.23]{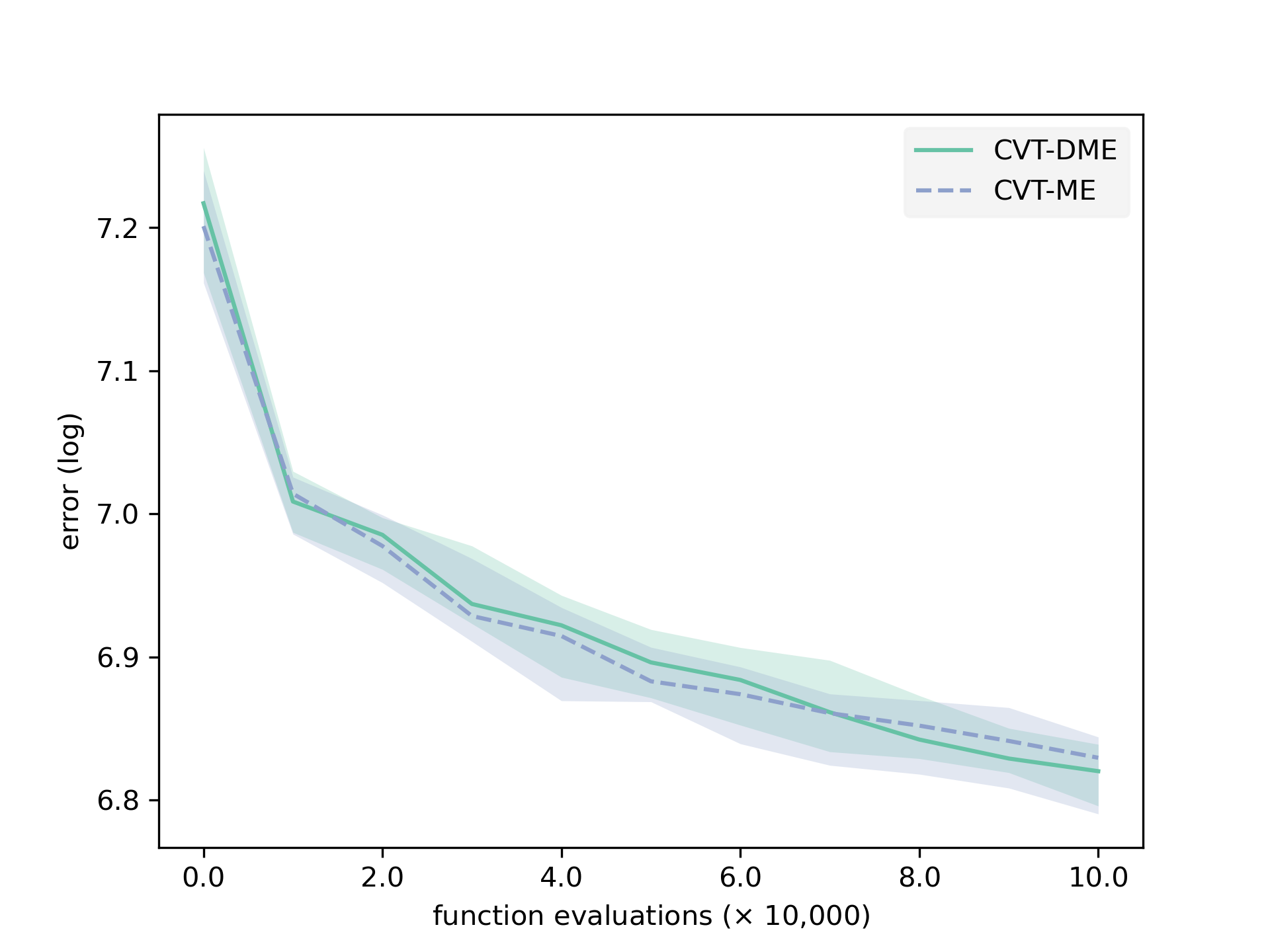}
   \label{fig:2_10_18}
   }
 \subfigure[$F_{20}$ ($n = 10$)]{
  \includegraphics[scale=0.23]{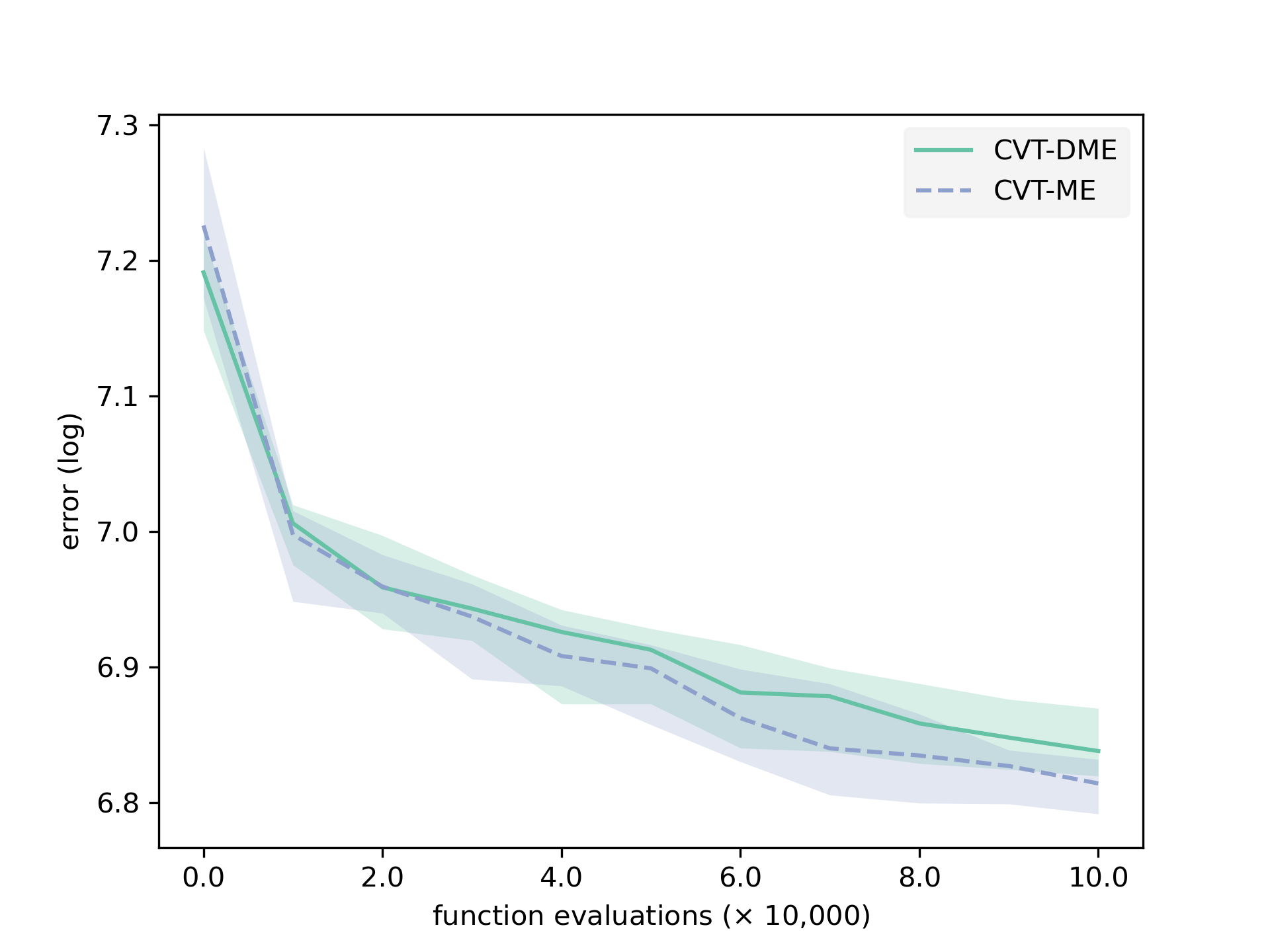}
   \label{fig:2_10_20}
   }
 \subfigure[$F_{24}$ ($n = 10$)]{
  \includegraphics[scale=0.23]{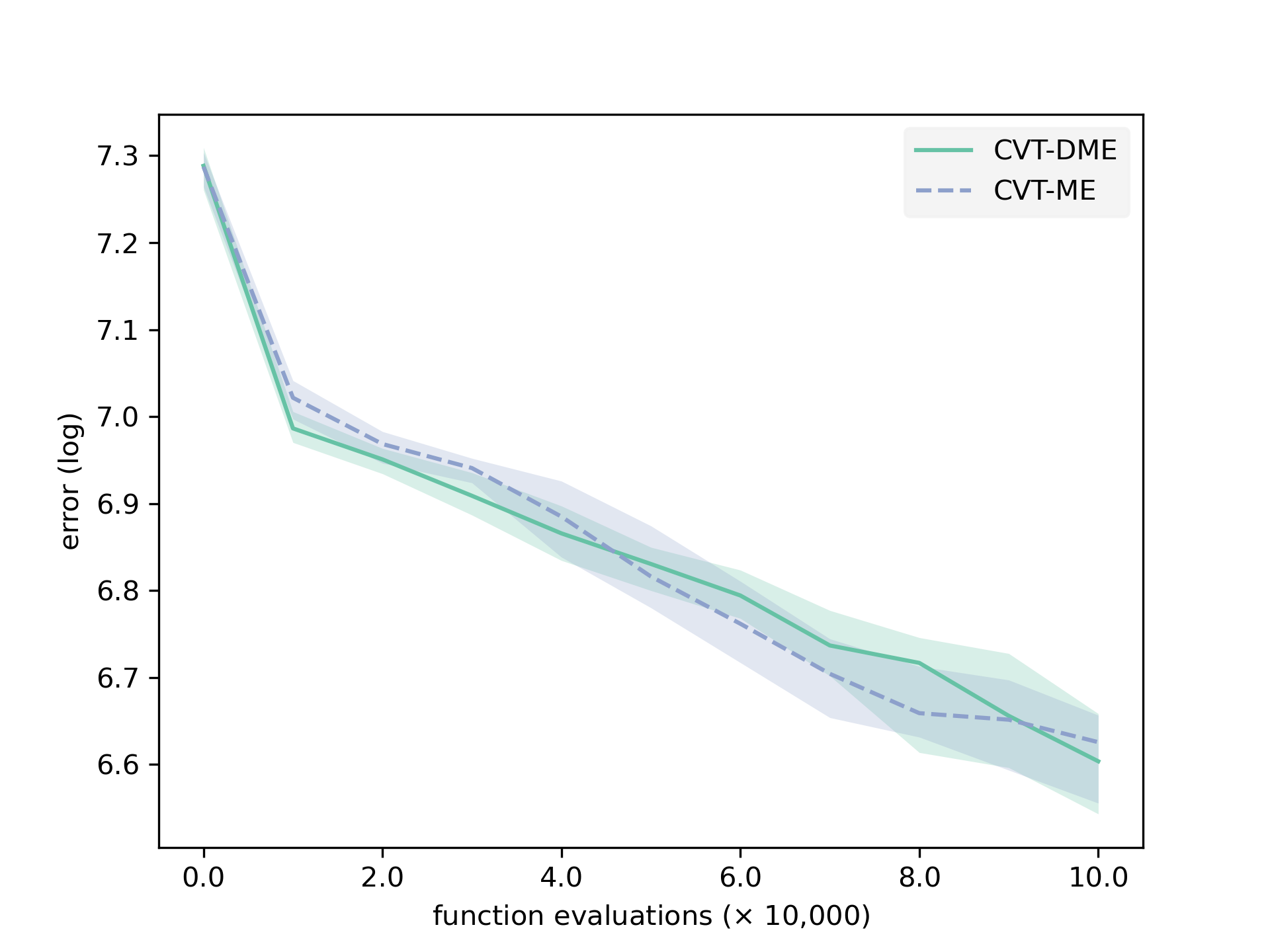}
   \label{fig:2_10_24}
   }
 \caption[]{Median and interquartile ranges (25th and 75th) of function error values ($n=10$)}
 \label{fig:medians1}
\end{figure*}

\begin{figure*}[pt]
 \centering
 \subfigure[CVT-DME on $F_{2}$ ($n = 10$)]{
  \includegraphics[scale=0.15]{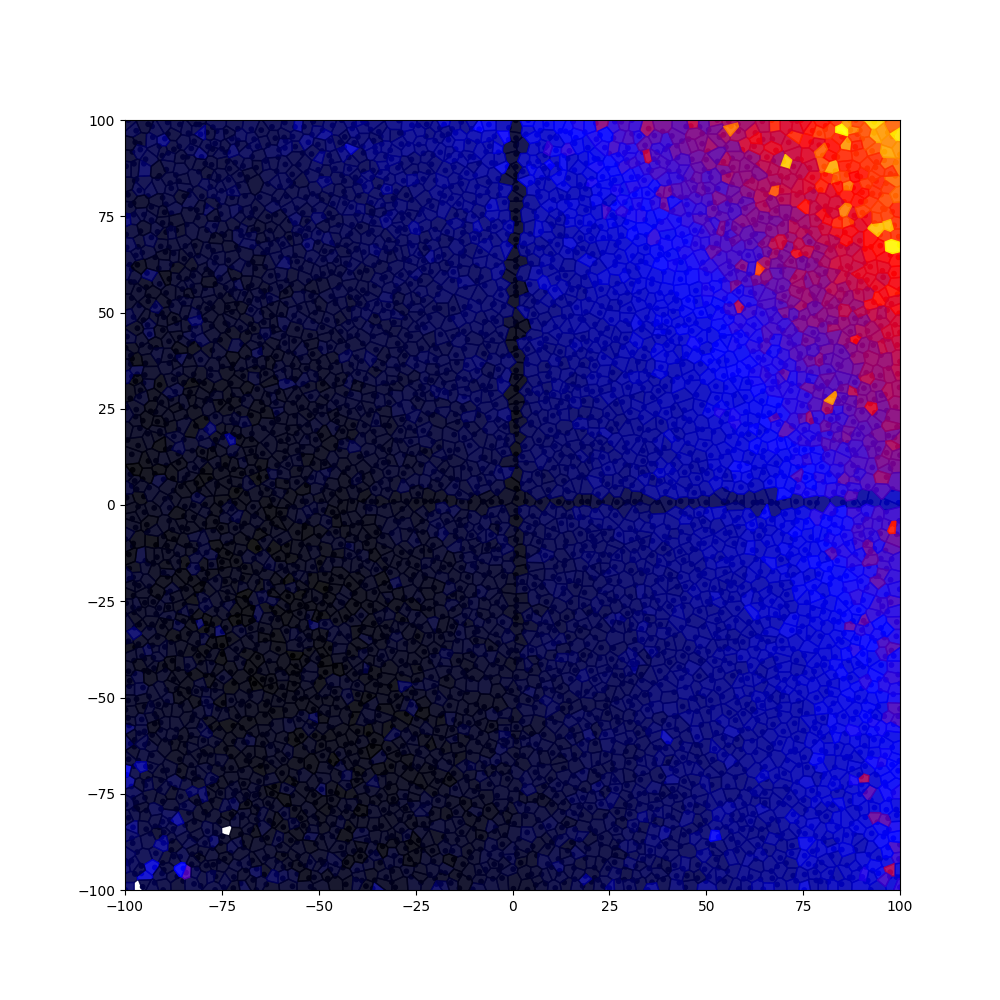}
   \label{fig:CVT-DME_2_10_2}
   }
 \subfigure[CVT-ME on $F_{2}$ ($n = 10$)]{
  \includegraphics[scale=0.15]{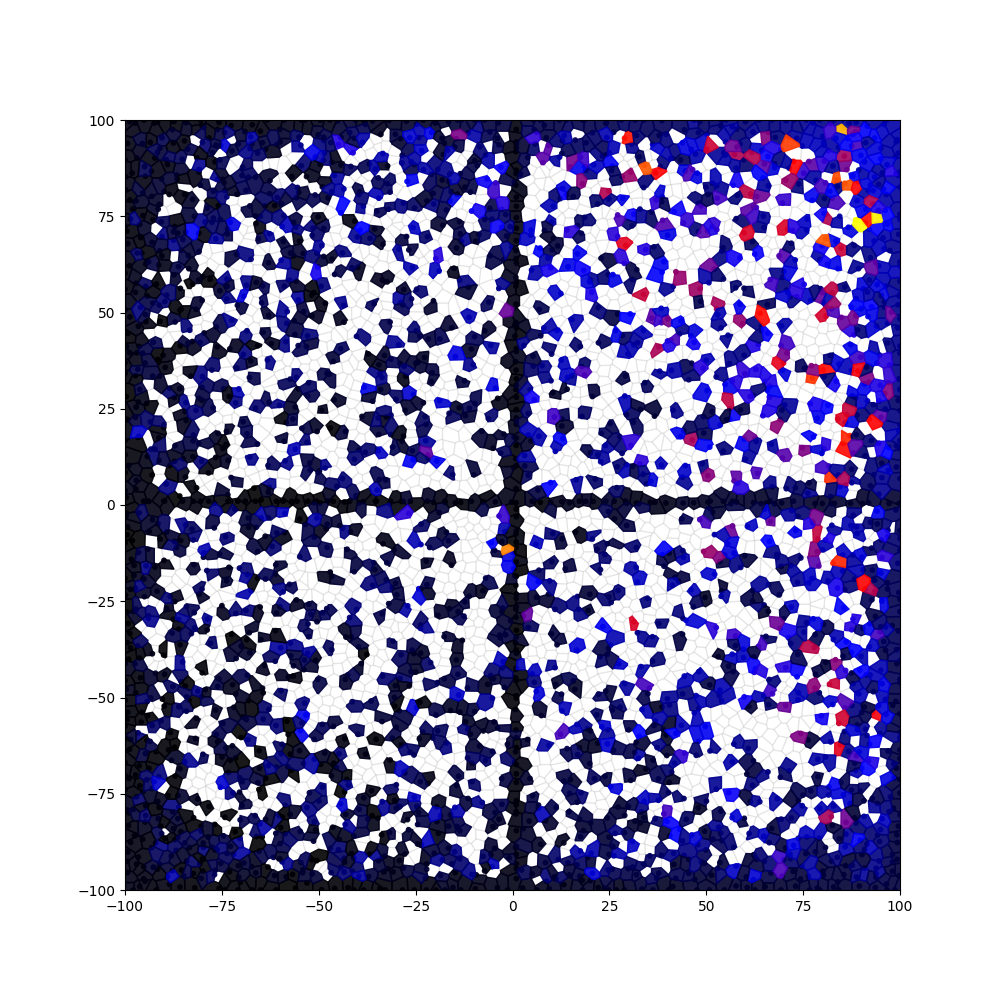}
   \label{fig:CVT-ME_2_10_2}
   }
 \subfigure[CVT-DME on $F_{14}$ ($n = 10$)]{
  \includegraphics[scale=0.15]{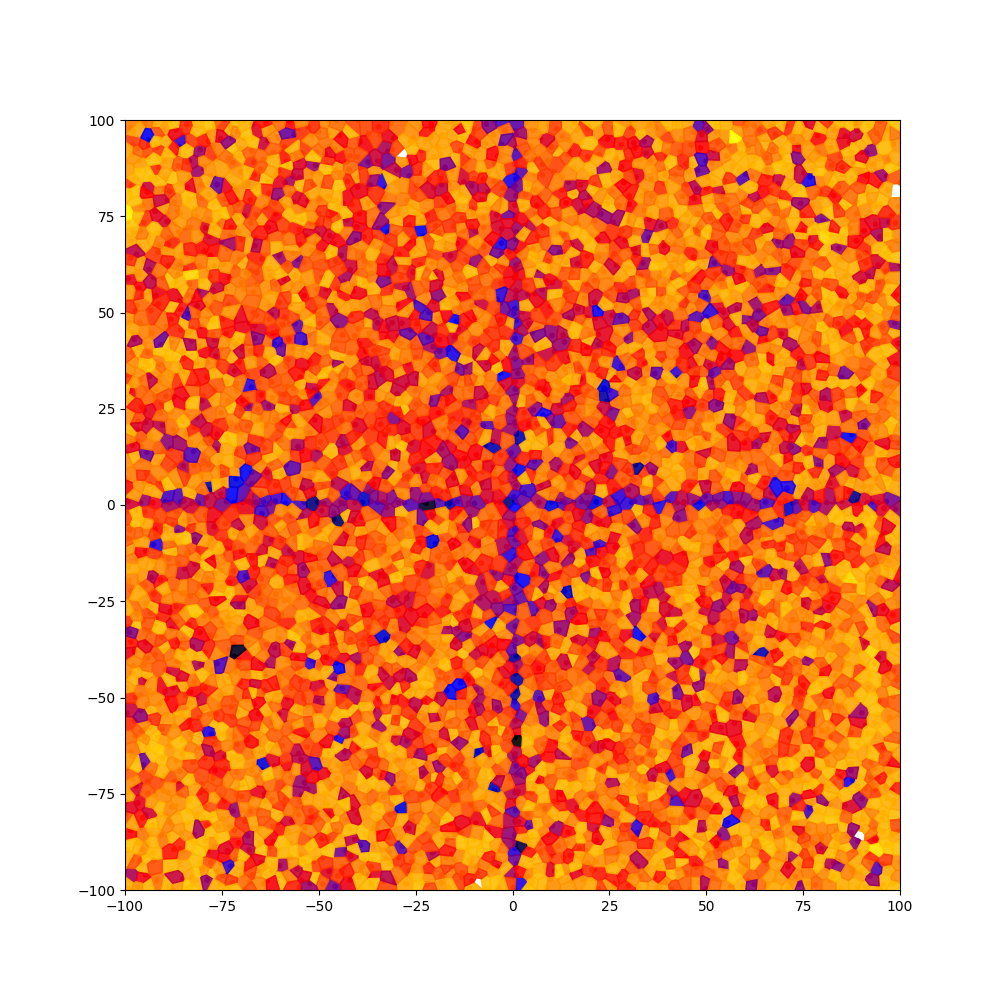}
   \label{fig:CVT-DME_2_10_14}
   }
 \subfigure[CVT-ME on $F_{14}$ ($n = 10$)]{
  \includegraphics[scale=0.15]{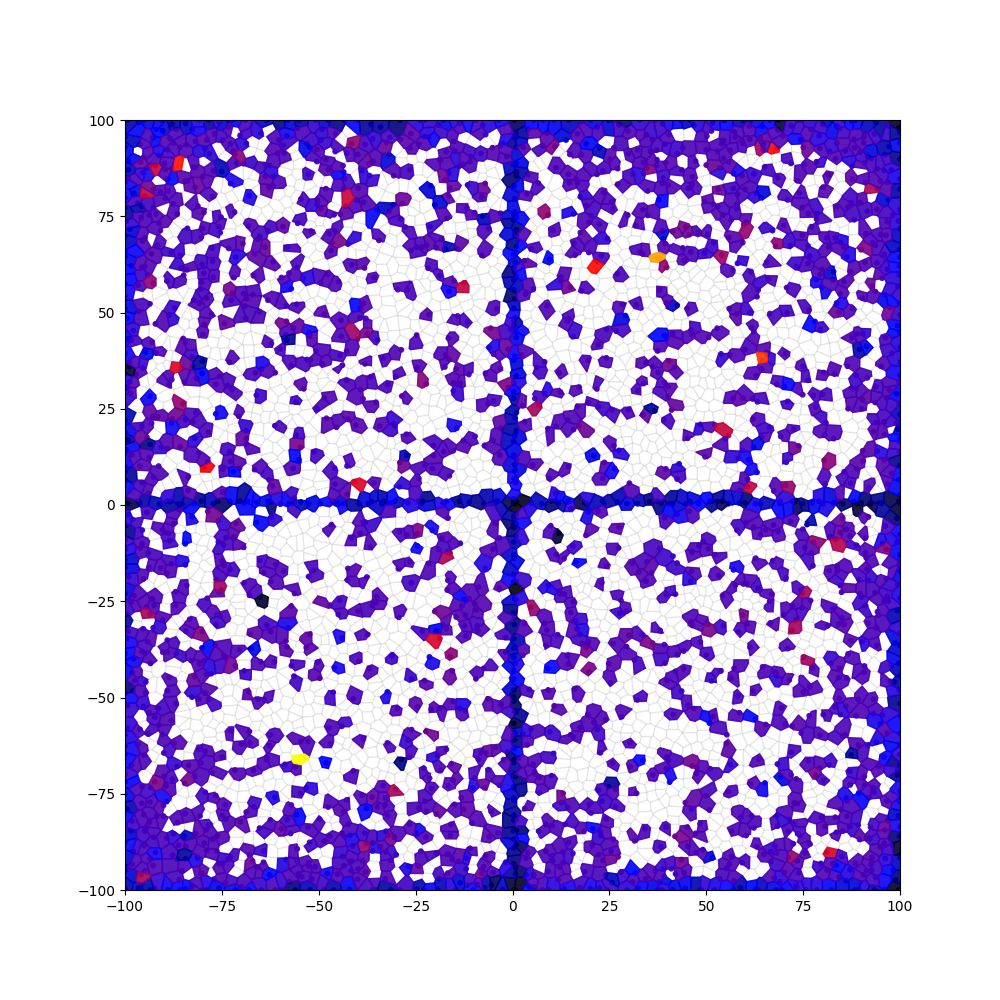}
   \label{fig:CVT-ME_2_10_14}
   }
 \caption[]{Final heatmaps found in a single run of CVT-DME and CVT-ME on $F_{2}$ and $F_{14}$ ($n=10$)}
 \label{fig:maps1}
\end{figure*}

Fig. \ref{fig:medians1} presents the convergence graphs of FEVs obtained by CVT-DME and CVT-ME. As we can see from the convergence graphs, the convergence speed of CVT-DME is faster than that of CVT-ME. These results emphasize that the vector difference based operators that CVT-DME uses are fast and effective in finding more accurate solutions compared to the Gaussian distribution based operators that CVT-ME uses. Also, Fig. \ref{fig:maps1} presents the final heatmaps obtained by CVT-DME and CVT-ME. As we can see from the heatmaps, CVT-DME fills more behavior space than CVT-ME does.

\begin{figure*}[pt]
 \centering
 \subfigure[$F_{2}$ ($n = 50$)]{
  \includegraphics[scale=0.23]{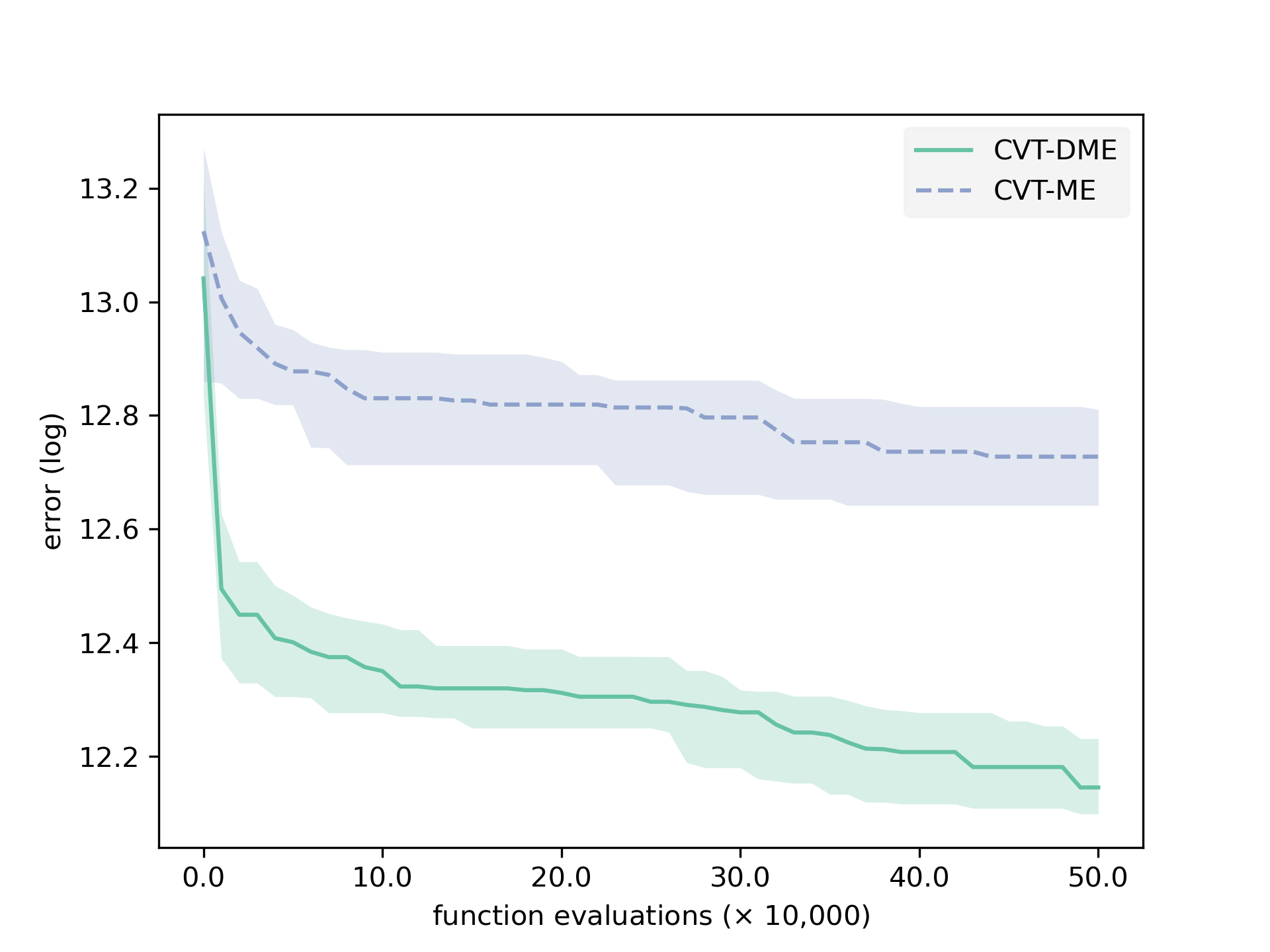}
   \label{fig:2_50_2}
   }
 \subfigure[$F_{4}$ ($n = 50$)]{
  \includegraphics[scale=0.23]{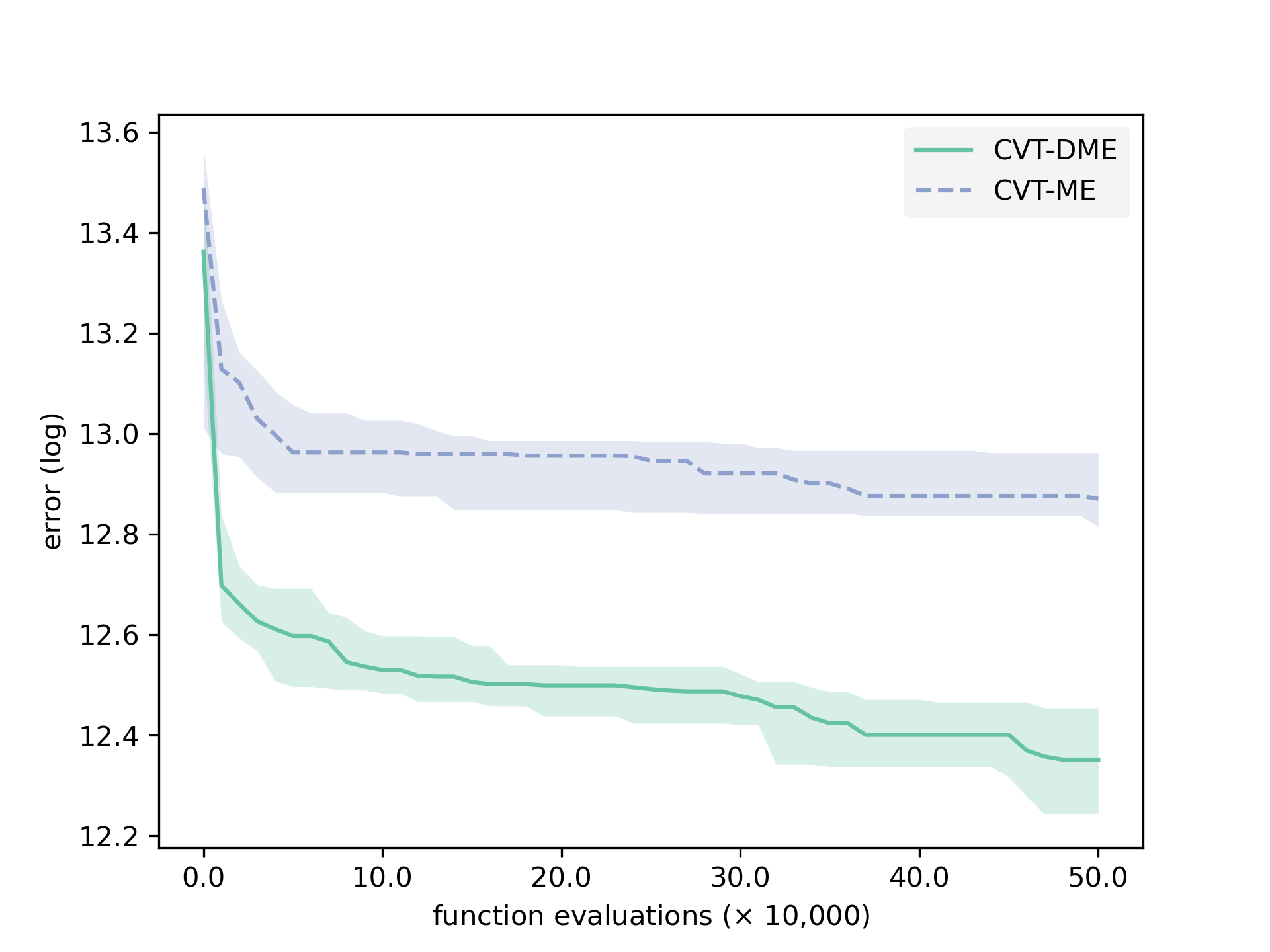}
   \label{fig:2_50_4}
   }
 \subfigure[$F_{6}$ ($n = 50$)]{
  \includegraphics[scale=0.23]{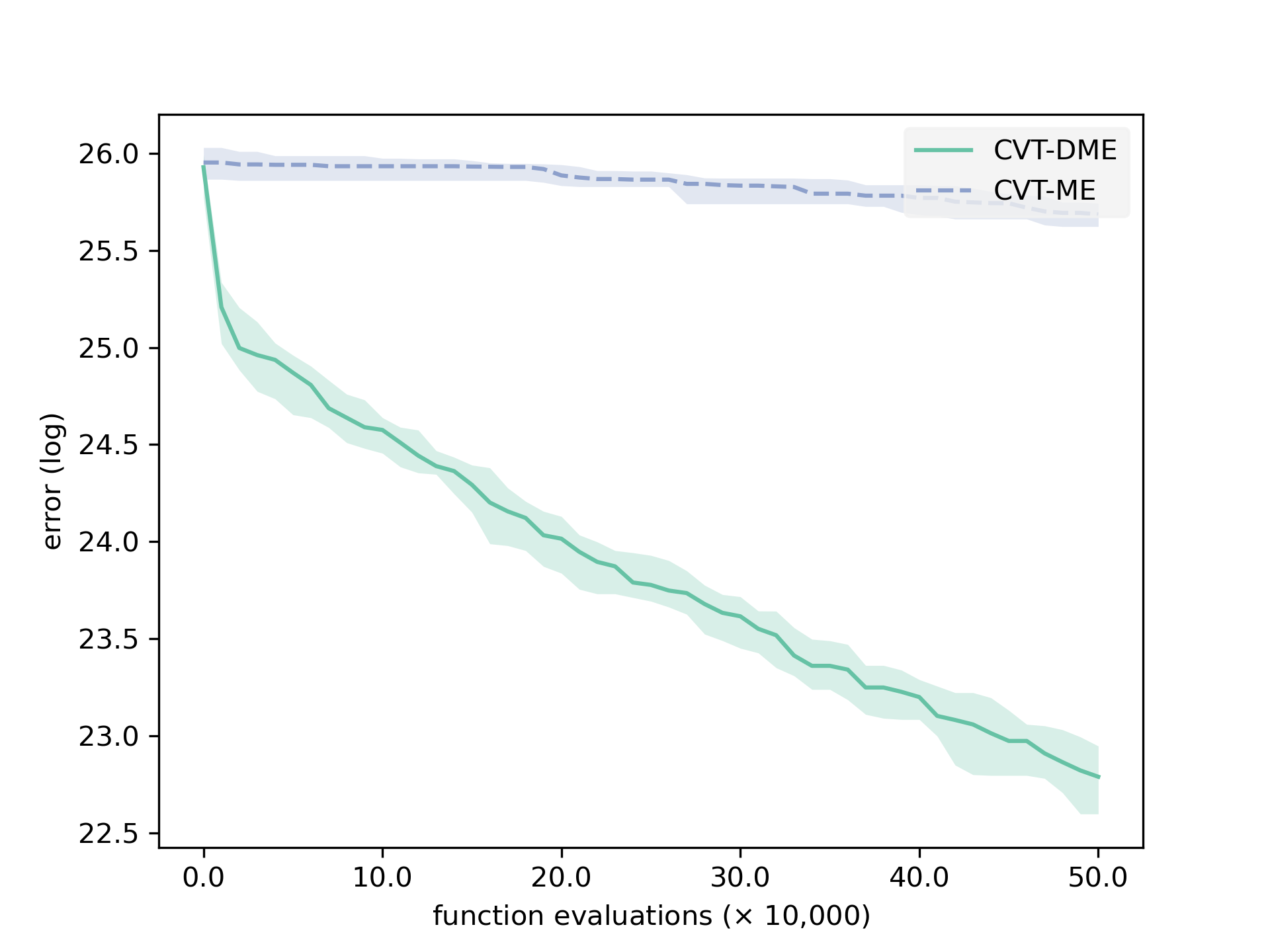}
   \label{fig:2_50_6}
   }
 \subfigure[$F_{8}$ ($n = 50$)]{
  \includegraphics[scale=0.23]{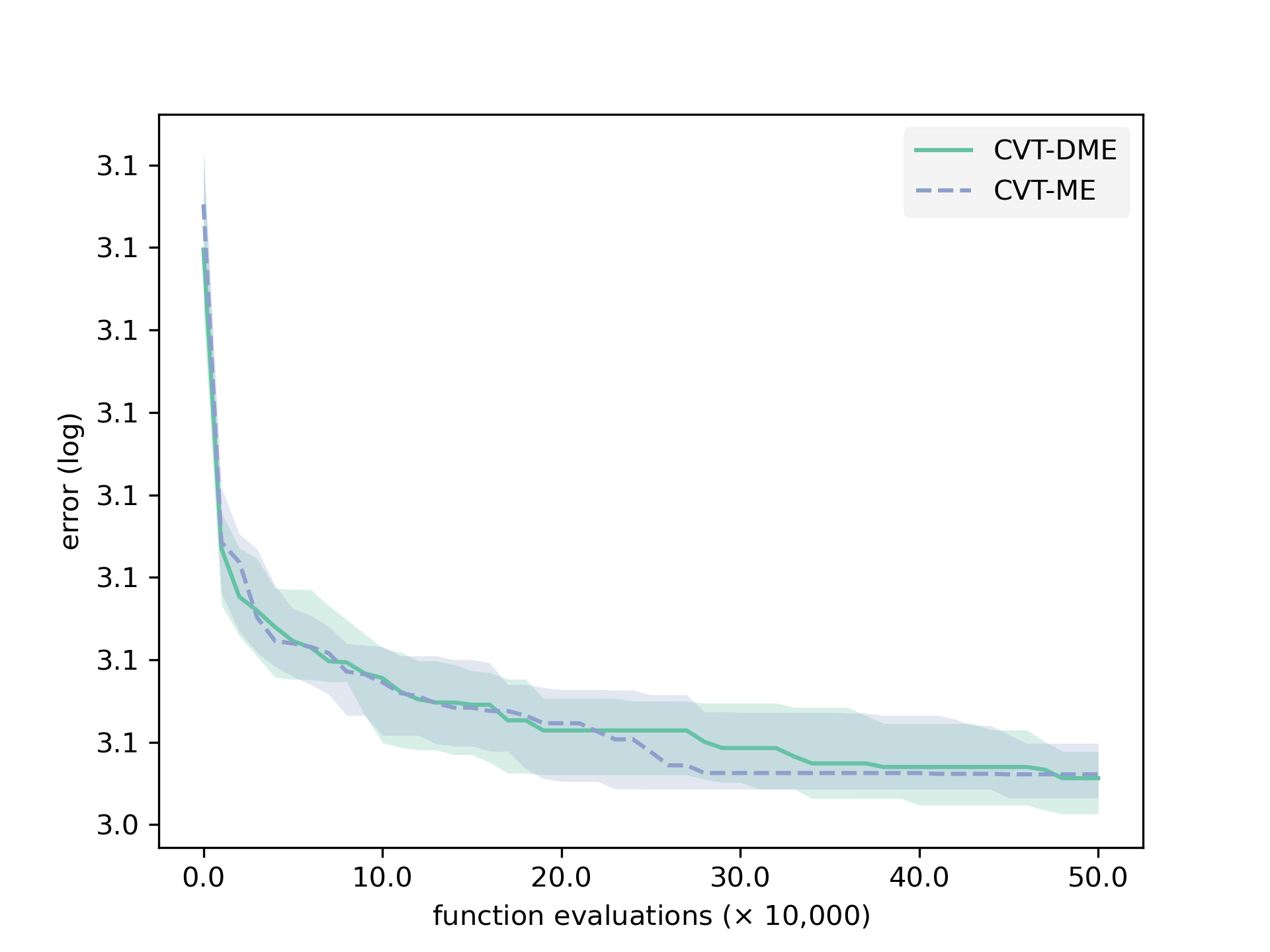}
   \label{fig:2_50_8}
   }
\subfigure[$F_{14}$ ($n = 50$)]{
  \includegraphics[scale=0.23]{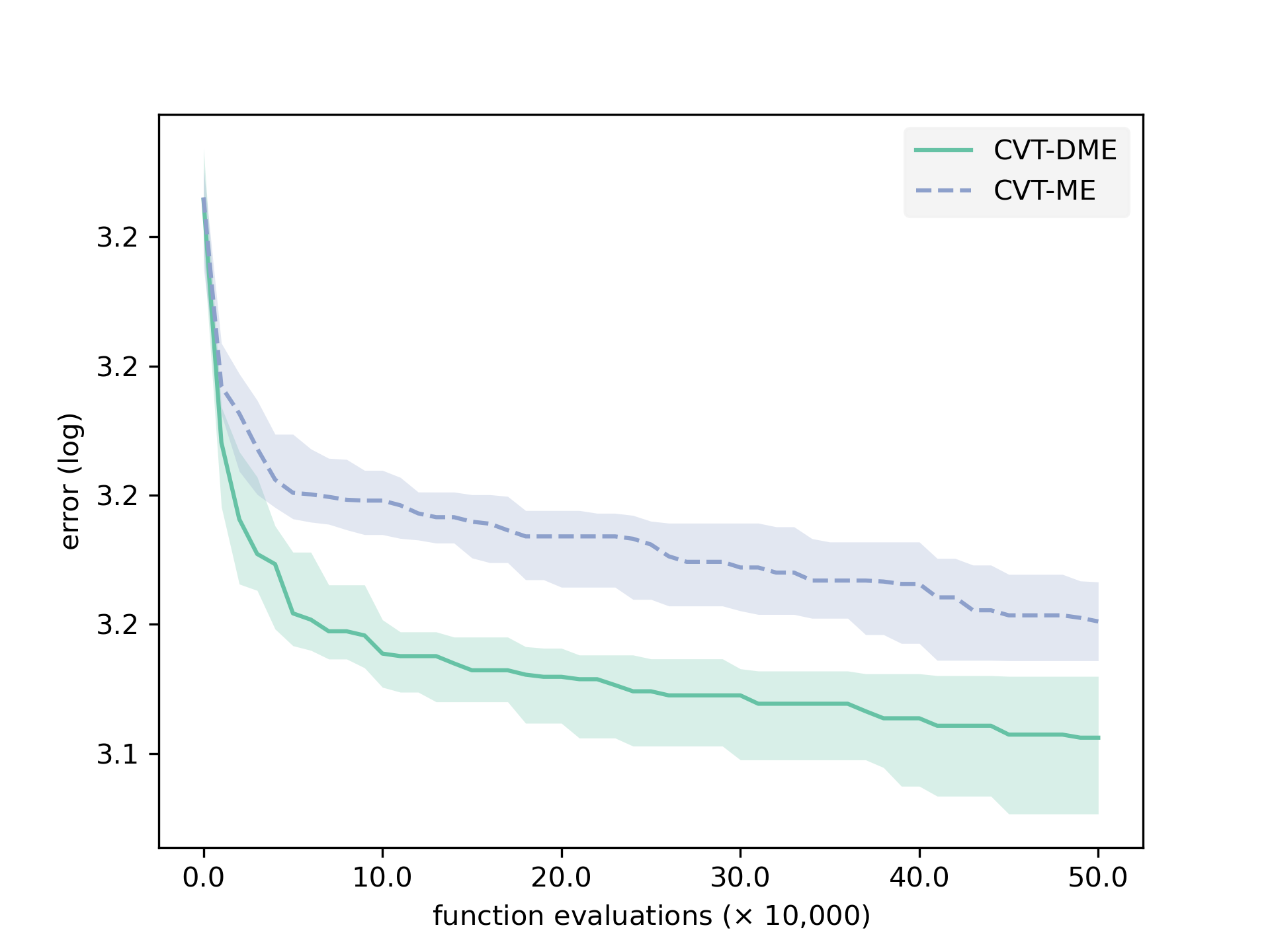}
   \label{fig:2_50_14}
   }
 \subfigure[$F_{18}$ ($n = 50$)]{
  \includegraphics[scale=0.23]{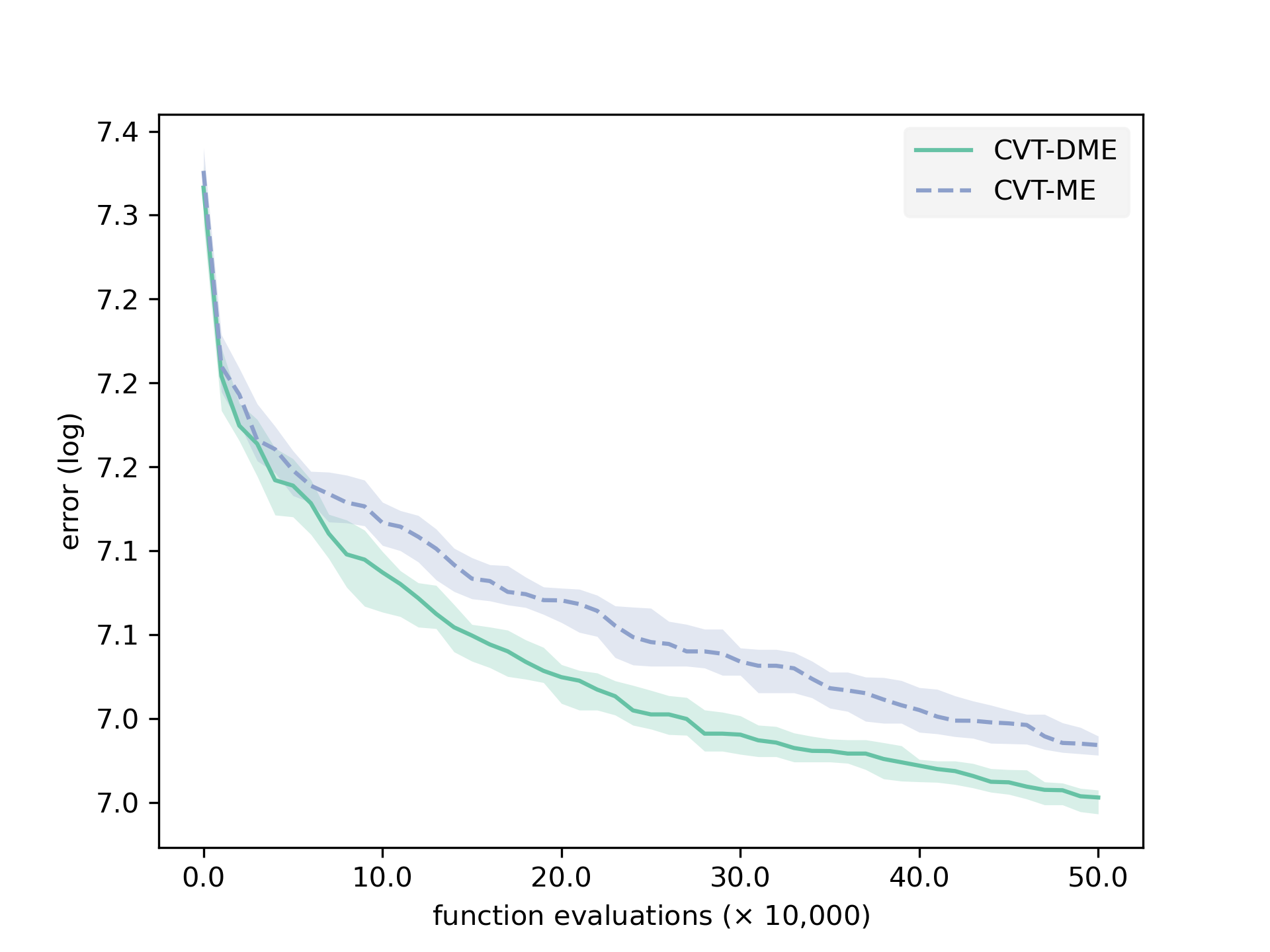}
   \label{fig:2_50_18}
   }
 \subfigure[$F_{20}$ ($n = 50$)]{
  \includegraphics[scale=0.23]{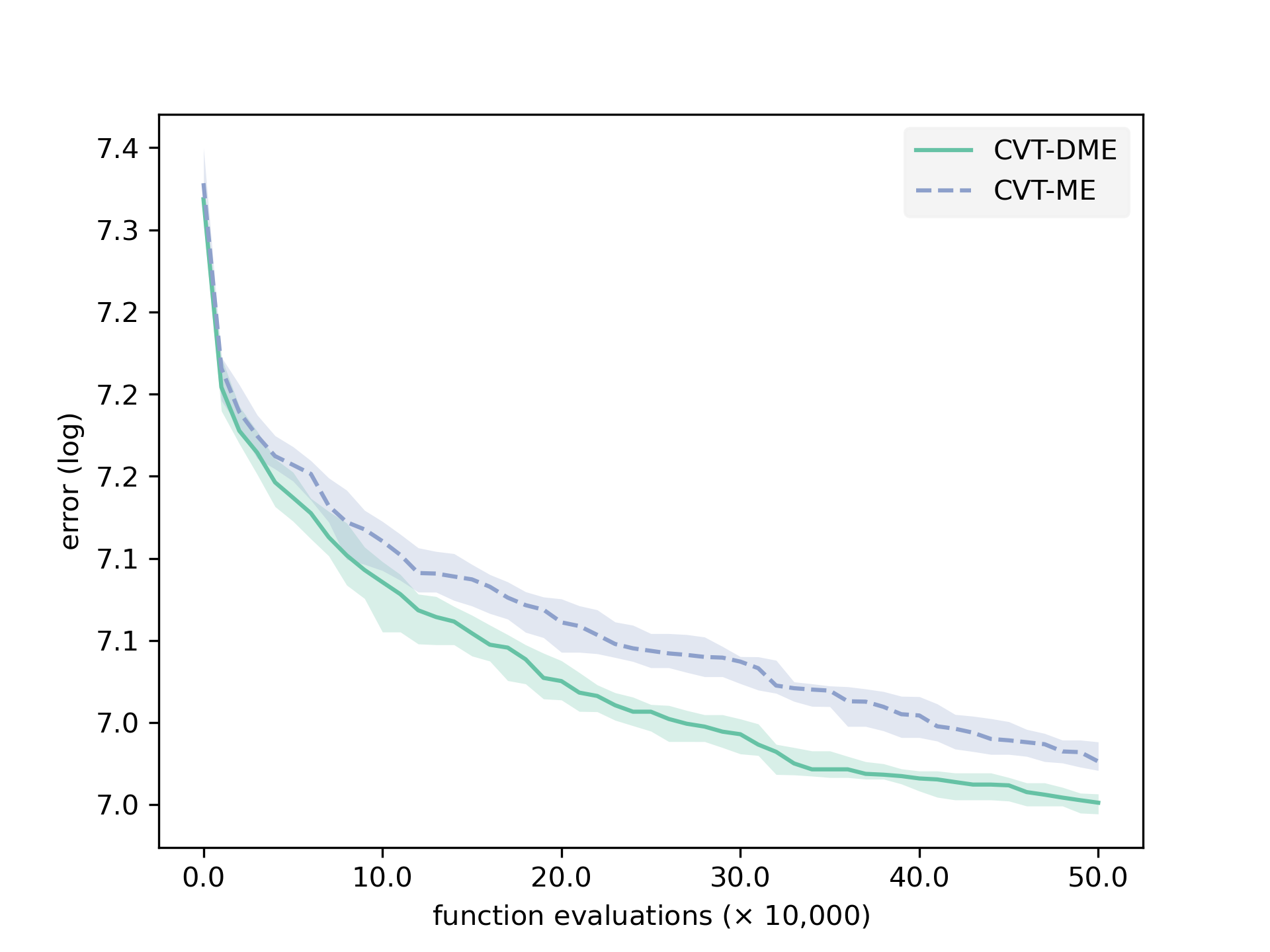}
   \label{fig:2_50_20}
   }
 \subfigure[$F_{24}$ ($n = 50$)]{
  \includegraphics[scale=0.23]{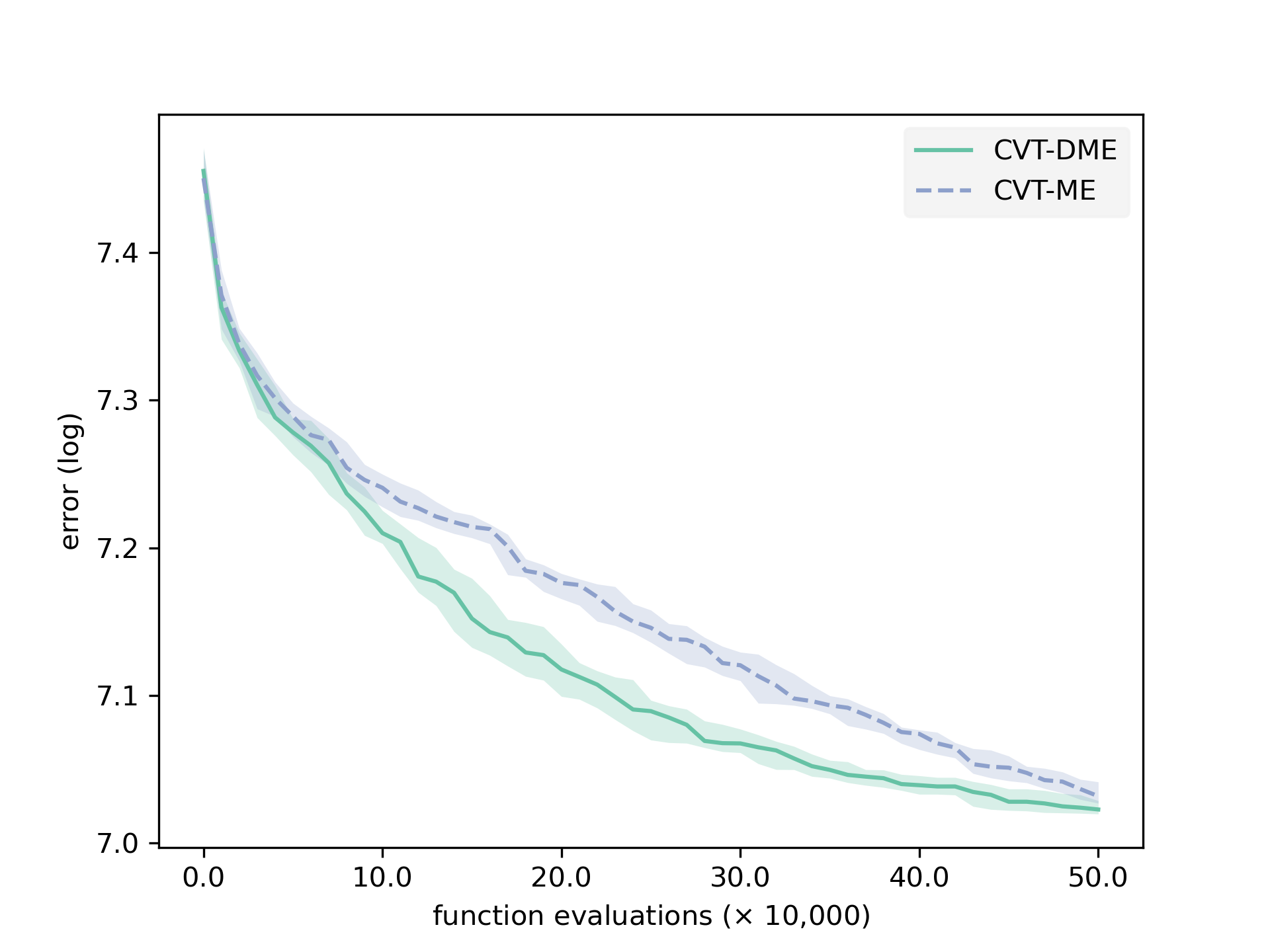}
   \label{fig:2_50_24}
   }
 \caption[]{Median and interquartile ranges (25th and 75th) of function error values ($n=50$)}
 \label{fig:medians2}
\end{figure*}

\begin{figure*}[pt]
 \centering
 \subfigure[CVT-DME on $F_{2}$ ($n = 50$)]{
  \includegraphics[scale=0.15]{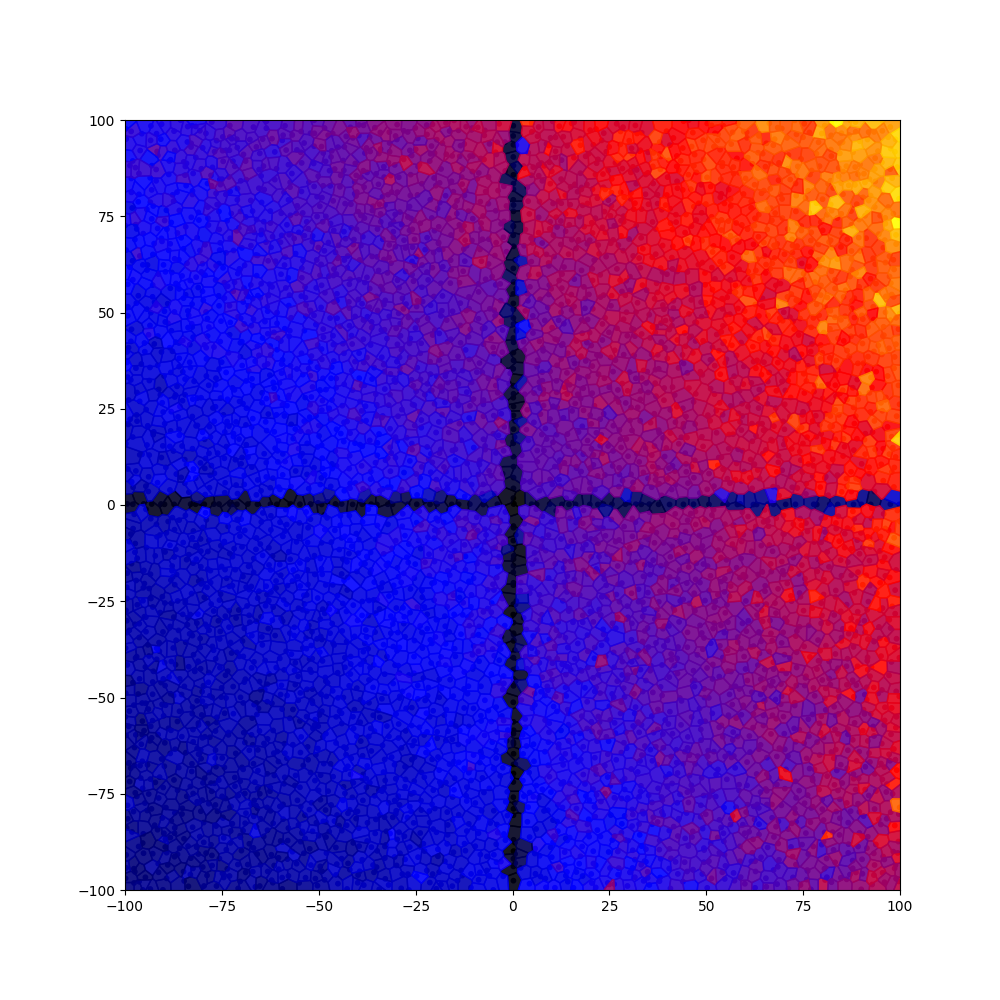}
   \label{fig:CVT-DME_2_50_2}
   }
 \subfigure[CVT-ME on $F_{2}$ ($n = 50$)]{
  \includegraphics[scale=0.15]{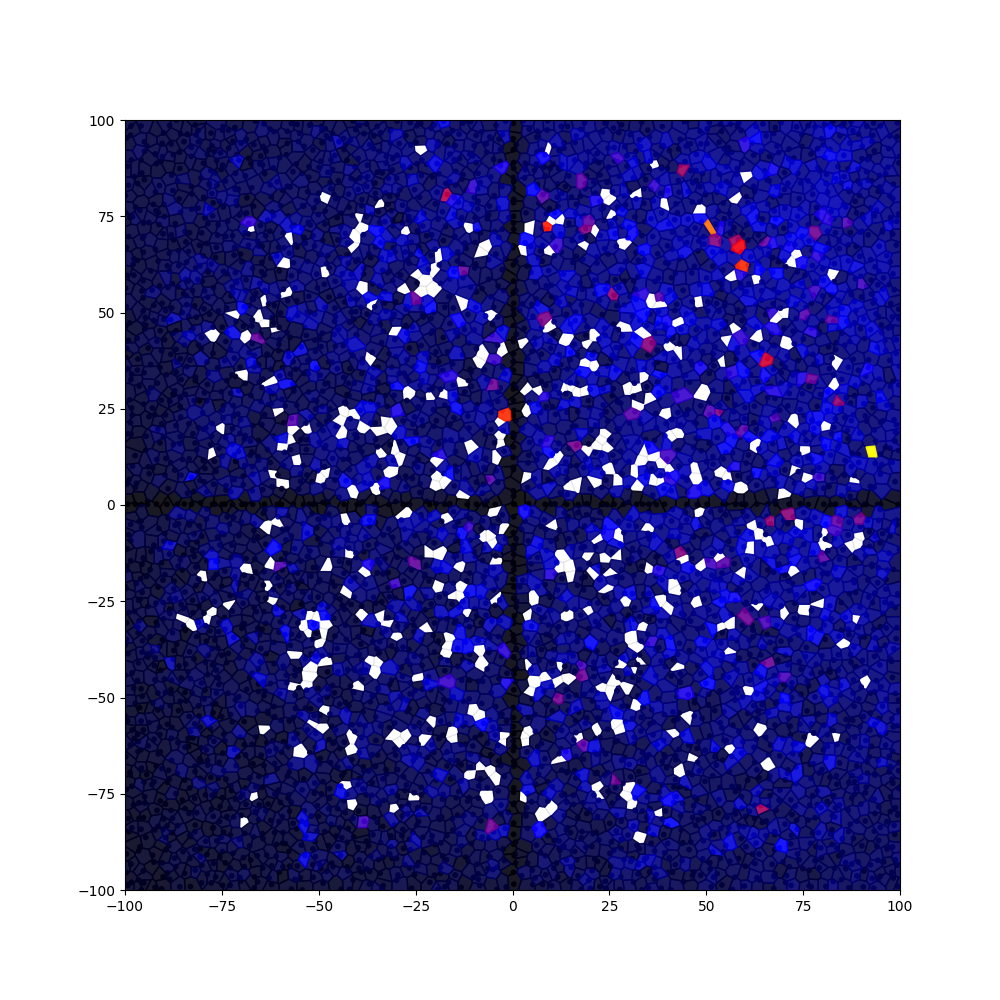}
   \label{fig:CVT-ME_2_50_2}
   }
 \subfigure[CVT-DME on $F_{14}$ ($n = 50$)]{
  \includegraphics[scale=0.15]{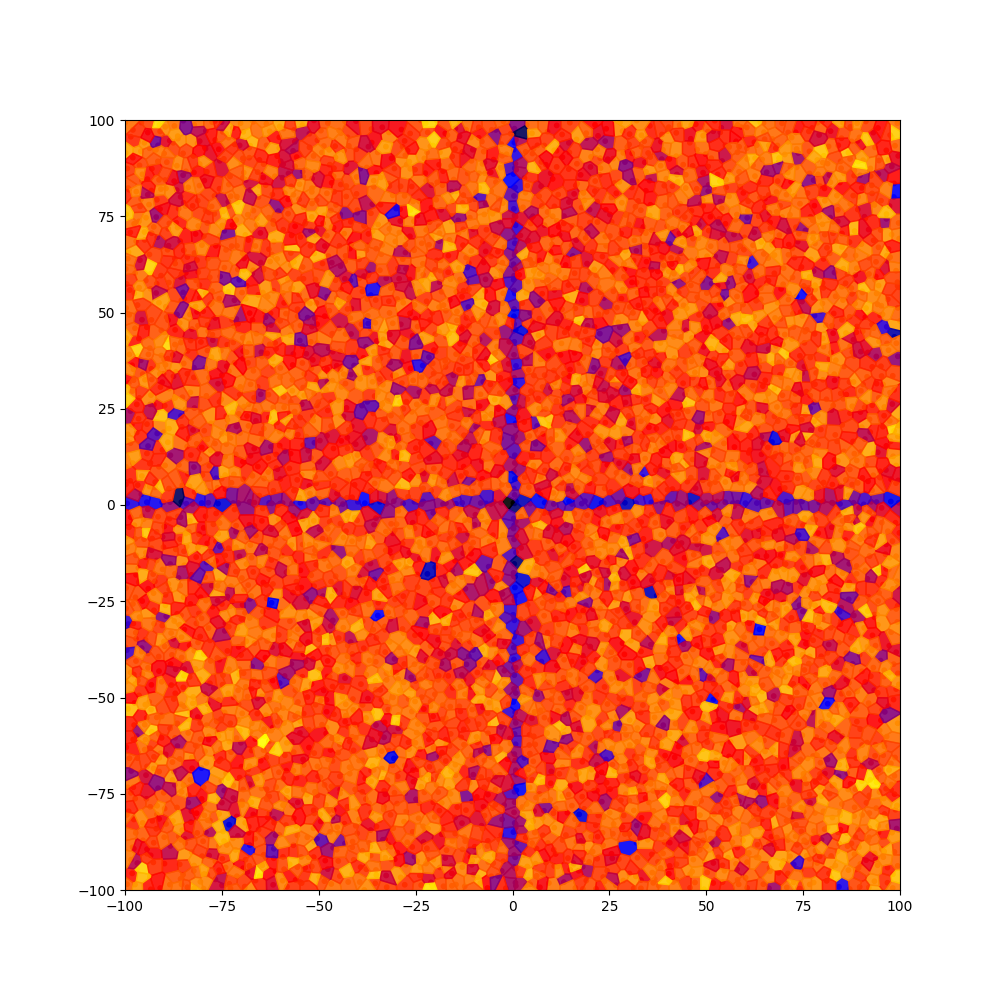}
   \label{fig:CVT-DME_2_50_14}
   }
 \subfigure[CVT-ME on $F_{14}$ ($n = 50$)]{
  \includegraphics[scale=0.15]{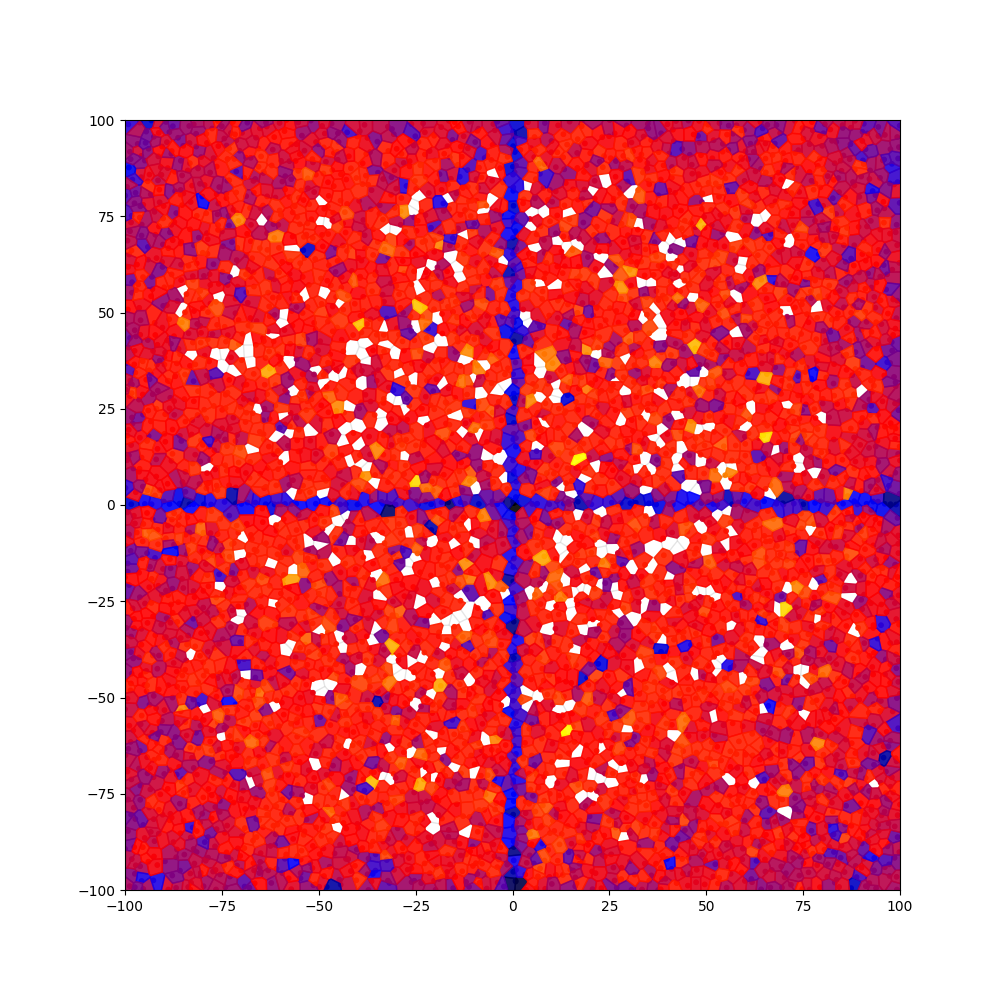}
   \label{fig:CVT-ME_2_50_14}
   }
 \caption[]{Final heatmaps found in a single run of CVT-DME and CVT-ME on $F_{2}$ and $F_{14}$ ($n=50$)}
 \label{fig:maps2}
\end{figure*}

We can see a similar tendency from Table \ref{tab:result2} and Figs. \ref{fig:medians2} and \ref{fig:maps2}. Note that the performance difference between CVT-DME and CVT-ME is greater as the dimension of a search space increases. Overall, the proposed algorithm Differential MAP-Elites finds not only more accurate solutions but also more diverse solutions compared to the comparison algorithm CVT-MAP-Elites.


\begin{table*}[htbp]
  \tiny
  \centering
  \caption{Experimental results of CVT-DME and CVT-ME in 30 independent runs on 25 test functions ($n = 2$ and $n = 10$)}
    \begin{tabular}{ccccc|cccc}
    \toprule
          & n=2  &       &       &       & n=10 &       &       &  \\
    \cmidrule(l){2-2} \cmidrule(l){6-6}
          & CVT-DME &       & CVT-ME &       & CVT-DME &       & CVT-ME &  \\
    \cmidrule(l){2-3} \cmidrule(l){4-5} \cmidrule(l){6-7} \cmidrule(l){8-9}
          & Mean (Std. Dev.) & Coverage (Std. Dev.) & Mean (Std. Dev.) & Coverage (Std. Dev.) & Mean (Std. Dev.) & Coverage (Std. Dev.) & Mean (Std. Dev.) & Coverage (Std. Dev.) \\
    \midrule
    F1    & \textbf{8.43E-01 (7.67E-01)} & \textbf{92.4\% (0.4\%)} & 2.17E+01 (1.96E+01) - & 15.7\% (0.5\%) - & \textbf{8.40E+02 (1.91E+02)} & \textbf{99.9\% (0.0\%)} & 1.54E+04 (2.70E+03) - & 60.0\% (0.9\%) - \\
    F2    & \textbf{1.12E+00 (1.02E+00)} & \textbf{92.4\% (0.4\%)} & 3.21E+01 (3.15E+01) - & 15.5\% (0.4\%) - & \textbf{1.06E+03 (2.92E+02)} & \textbf{99.9\% (0.1\%)} & 1.07E+04 (1.51E+03) - & 52.7\% (1.4\%) - \\
    F3    & \textbf{8.15E+02 (8.18E+02)} & \textbf{92.5\% (0.3\%)} & 6.50E+03 (5.24E+03) - & 15.5\% (0.5\%) - & \textbf{5.48E+06 (1.92E+06)} & \textbf{100.0\% (0.0\%)} & 4.57E+07 (1.86E+07) - & 47.7\% (1.1\%) - \\
    F4    & \textbf{1.36E+00 (1.78E+00)} & \textbf{92.4\% (0.4\%)} & 2.53E+01 (2.46E+01) - & 15.3\% (0.5\%) - & \textbf{1.48E+03 (4.38E+02)} & \textbf{99.9\% (0.1\%)} & 1.30E+04 (1.71E+03) - & 51.9\% (1.5\%) - \\
    F5    & 0.00E+00 (0.00E+00) & \textbf{92.3\% (0.3\%)} & 0.00E+00 (0.00E+00) = & 15.4\% (0.4\%) - & 6.81E+03 (7.74E+02) & \textbf{99.9\% (0.1\%)} & \textbf{5.38E+03 (6.38E+02) +} & 51.3\% (1.4\%) - \\
    F6    & \textbf{1.08E+01 (8.72E+00)} & \textbf{92.3\% (0.4\%)} & 1.25E+02 (9.98E+01) - & 15.6\% (0.6\%) - & \textbf{8.06E+06 (3.72E+06)} & \textbf{99.9\% (0.1\%)} & 1.97E+09 (7.63E+08) - & 61.7\% (0.9\%) - \\
    F7    & 3.94E+01 (1.33E-01) & 0.0\% (0.0\%) & \textbf{3.93E+01 (2.17E-14) +} & 0.0\% (0.0\%) = & 1.51E+03 (6.31E+02) & 0.0\% (0.0\%) & 1.27E+03 (0.00E+00) = & 0.0\% (0.0\%) = \\
    F8    & 2.78E+00 (1.74E+00) & \textbf{92.4\% (0.4\%)} & \textbf{1.53E+00 (1.52E+00) +} & 46.1\% (0.7\%) - & 2.04E+01 (9.00E-02) & \textbf{99.9\% (0.1\%)} & 2.03E+01 (7.28E-02) = & 76.9\% (0.6\%) - \\
    F9    & 3.01E-01 (2.77E-01) & \textbf{92.4\% (0.3\%)} & 2.57E-01 (2.68E-01) = & 90.9\% (0.4\%) - & 4.74E+01 (5.26E+00) & 99.9\% (0.0\%) & 4.44E+01 (6.01E+00) = & 99.9\% (0.1\%) = \\
    F10   & 5.43E-01 (4.49E-01) & \textbf{92.4\% (0.4\%)} & 5.22E-01 (4.05E-01) = & 90.9\% (0.4\%) - & 5.78E+01 (5.21E+00) & 99.9\% (0.0\%) & 5.72E+01 (6.81E+00) = & 99.9\% (0.1\%) = \\
    F11   & 2.79E-01 (1.03E-01) & 92.4\% (0.3\%) & 2.25E-01 (8.37E-02) = & \textbf{93.1\% (0.7\%) +} & 9.09E+00 (6.02E-01) & 99.9\% (0.1\%) & \textbf{6.00E+00 (7.06E-01) +} & \textbf{100.0\% (0.0\%) +} \\
    F12   & 1.20E+00 (1.32E+00) & \textbf{92.6\% (0.3\%)} & 9.74E-01 (8.01E-01) = & 92.4\% (0.3\%) - & 1.92E+04 (4.92E+03) & 99.9\% (0.0\%) & \textbf{7.17E+03 (1.85E+03) +} & 99.9\% (0.1\%) = \\
    F13   & 3.35E-02 (3.76E-02) & \textbf{92.2\% (0.3\%)} & 2.99E-02 (3.68E-02) = & 90.7\% (0.4\%) - & 4.03E+01 (2.13E+01) & \textbf{99.9\% (0.1\%)} & \textbf{1.70E+01 (5.59E+00) +} & 99.9\% (0.1\%) - \\
    F14   & \textbf{6.00E-02 (3.40E-02)} & \textbf{92.3\% (0.4\%)} & 3.63E-01 (2.20E-01) - & 15.5\% (0.5\%) - & \textbf{3.86E+00 (1.56E-01)} & \textbf{99.9\% (0.1\%)} & 4.11E+00 (7.06E-02) - & 44.4\% (0.8\%) - \\
    F15   & 1.94E+01 (1.98E+01) & \textbf{92.3\% (0.3\%)} & 1.59E+01 (1.45E+01) = & 90.9\% (0.4\%) - & 5.17E+02 (5.25E+01) & \textbf{99.9\% (0.1\%)} & 5.23E+02 (4.32E+01) = & 99.9\% (0.1\%) - \\
    F16   & \textbf{1.14E+01 (1.21E+01)} & \textbf{92.4\% (0.3\%)} & 2.22E+01 (1.46E+01) - & 90.7\% (0.3\%) - & 2.33E+02 (1.87E+01) & 99.9\% (0.1\%) & \textbf{2.24E+02 (1.05E+01) +} & 99.9\% (0.1\%) = \\
    F17   & 2.02E+01 (1.65E+01) & \textbf{92.5\% (0.4\%)} & 1.69E+01 (1.78E+01) = & 90.9\% (0.5\%) - & 2.59E+02 (1.80E+01) & 99.9\% (0.0\%) & \textbf{2.43E+02 (1.50E+01) +} & 99.9\% (0.0\%) = \\
    F18   & 1.21E+02 (6.19E+01) & \textbf{92.2\% (0.3\%)} & 1.19E+02 (5.90E+01) = & 90.8\% (0.4\%) - & 9.16E+02 (2.96E+01) & \textbf{99.9\% (0.1\%)} & 9.13E+02 (4.01E+01) = & 99.9\% (0.1\%) - \\
    F19   & 2.08E+02 (2.42E+01) & \textbf{92.4\% (0.3\%)} & 2.17E+02 (1.80E+01) = & 90.9\% (0.4\%) - & 9.14E+02 (4.76E+01) & 99.9\% (0.1\%) & 9.10E+02 (3.97E+01) = & 99.9\% (0.1\%) = \\
    F20   & 7.02E+00 (7.50E+00) & \textbf{92.2\% (0.3\%)} & 6.97E+00 (6.49E+00) = & 90.5\% (0.3\%) - & 9.32E+02 (3.00E+01) & 99.9\% (0.1\%) & \textbf{9.05E+02 (2.82E+01) +} & 99.9\% (0.0\%) = \\
    F21   & 8.83E+01 (7.37E+01) & \textbf{92.5\% (0.3\%)} & 1.09E+02 (7.92E+01) = & 90.9\% (0.4\%) - & 1.05E+03 (5.84E+01) & 99.9\% (0.0\%) & \textbf{1.02E+03 (5.62E+01) +} & 99.9\% (0.0\%) = \\
    F22   & 1.78E+02 (6.10E+01) & \textbf{92.4\% (0.3\%)} & 1.67E+02 (6.21E+01) = & 90.9\% (0.4\%) - & \textbf{8.74E+02 (1.59E+01)} & \textbf{99.9\% (0.1\%)} & 8.86E+02 (1.98E+01) - & 99.9\% (0.0\%) - \\
    F23   & 9.99E+01 (7.28E+01) & \textbf{92.4\% (0.3\%)} & 9.23E+01 (7.92E+01) = & 90.9\% (0.4\%) - & 1.06E+03 (6.30E+01) & 99.9\% (0.0\%) & 1.03E+03 (9.13E+01) = & 99.9\% (0.1\%) = \\
    F24   & 1.95E+02 (1.74E+01) & \textbf{92.5\% (0.3\%)} & 1.97E+02 (1.60E+01) = & 91.0\% (0.4\%) - & 7.39E+02 (6.37E+01) & \textbf{100.0\% (0.0\%)} & 7.37E+02 (6.49E+01) = & 100.0\% (0.1\%) - \\
    F25   & 1.69E+02 (2.98E+01) & 0.0\% (0.0\%) & 1.59E+02 (1.99E+01) = & 0.0\% (0.0\%) = & 2.09E+03 (6.91E+01) & 0.0\% (0.0\%) & \textbf{2.03E+03 (6.20E+01) +} & 0.0\% (0.0\%) = \\
    \midrule
    +/=/- &       &       & 2/16/7 & 1/2/22 &       &       & 9/9/7 & 1/11/13 \\
    \bottomrule
    \end{tabular}%
  \label{tab:result1}%
\\The symbols "+/=/-" indicate that CVT-ME performed significantly better ($+$), not significantly better or worse ($=$), or significantly worse ($-$) compared to CVT-DME using the Wilcoxon rank sum test with $\alpha = 0.05$ significance level.
\end{table*}%

\begin{table*}[pt]
  \tiny
  \centering
  \caption{Experimental results of CVT-DME and CVT-ME in 30 independent runs on 25 test functions ($n = 30$ and $n = 50$)}
    \begin{tabular}{ccccc|cccc}
    \toprule
          & n=30  &       &       &       & n=50 &       &       &  \\
    \cmidrule(l){2-2} \cmidrule(l){6-6}
          & CVT-DME &       & CVT-ME &       & CVT-DME &       & CVT-ME &  \\
    \cmidrule(l){2-3} \cmidrule(l){4-5} \cmidrule(l){6-7} \cmidrule(l){8-9}
          & Mean (Std. Dev.) & Coverage (Std. Dev.) & Mean (Std. Dev.) & Coverage (Std. Dev.) & Mean (Std. Dev.) & Coverage (Std. Dev.) & Mean (Std. Dev.) & Coverage (Std. Dev.) \\
    \midrule
    F1    & \textbf{1.05E+04 (1.49E+03)} & \textbf{100.0\% (0.0\%)} & 8.59E+04 (6.05E+03) - & 89.8\% (0.6\%) - & \textbf{3.25E+04 (5.25E+03)} & \textbf{100.0\% (0.0\%)} & 1.81E+05 (1.16E+04) - & 95.2\% (0.4\%) - \\
    F2    & \textbf{4.31E+04 (4.85E+03)} & \textbf{100.0\% (0.0\%)} & 1.13E+05 (1.30E+04) - & 82.7\% (0.5\%) - & \textbf{1.88E+05 (2.21E+04)} & \textbf{100.0\% (0.0\%)} & 3.35E+05 (4.52E+04) - & 92.8\% (0.3\%) - \\
    F3    & \textbf{1.71E+08 (3.04E+07)} & \textbf{100.0\% (0.0\%)} & 6.96E+08 (1.33E+08) - & 81.7\% (0.6\%) - & \textbf{7.48E+08 (1.06E+08)} & \textbf{100.0\% (0.0\%)} & 2.48E+09 (3.62E+08) - & 93.3\% (0.4\%) - \\
    F4    & \textbf{5.22E+04 (6.28E+03)} & \textbf{100.0\% (0.0\%)} & 1.34E+05 (1.73E+04) - & 82.5\% (0.6\%) - & \textbf{2.33E+05 (2.77E+04)} & \textbf{100.0\% (0.0\%)} & 3.93E+05 (4.09E+04) - & 92.8\% (0.4\%) - \\
    F5    & \textbf{1.99E+04 (1.00E+03)} & \textbf{100.0\% (0.0\%)} & 2.62E+04 (1.35E+03) - & 85.6\% (0.6\%) - & \textbf{3.12E+04 (1.38E+03)} & \textbf{100.0\% (0.0\%)} & 4.79E+04 (1.99E+03) - & 92.7\% (0.5\%) - \\
    F6    & \textbf{1.77E+09 (3.88E+08)} & \textbf{100.0\% (0.0\%)} & 4.55E+10 (3.70E+09) - & 88.8\% (0.7\%) - & \textbf{7.65E+09 (1.99E+09)} & \textbf{100.0\% (0.0\%)} & 1.43E+11 (1.46E+10) - & 94.8\% (0.4\%) - \\
    F7    & 1.19E+04 (7.28E+02) & 0.0\% (0.0\%) & \textbf{4.77E+03 (1.98E+01) +} & 0.0\% (0.0\%) = & 1.67E+04 (5.84E+02) & 0.0\% (0.0\%) & \textbf{6.25E+03 (1.59E+01) +} & 0.0\% (0.0\%) = \\
    F8    & 2.09E+01 (6.61E-02) & \textbf{100.0\% (0.0\%)} & 2.09E+01 (7.30E-02) = & 94.7\% (0.3\%) - & 2.11E+01 (5.35E-02) & \textbf{100.0\% (0.0\%)} & 2.11E+01 (5.83E-02) = & 97.7\% (0.2\%) - \\
    F9    & 2.45E+02 (9.12E+00) & 100.0\% (0.0\%) & 2.42E+02 (1.14E+01) = & 100.0\% (0.0\%) = & 5.11E+02 (1.61E+01) & 100.0\% (0.0\%) & \textbf{4.99E+02 (1.76E+01) +} & 100.0\% (0.0\%) = \\
    F10   & 3.04E+02 (1.31E+01) & 100.0\% (0.0\%) & 3.01E+02 (1.49E+01) = & 100.0\% (0.0\%) = & 6.48E+02 (3.47E+01) & 100.0\% (0.0\%) & 6.46E+02 (2.23E+01) = & 100.0\% (0.0\%) = \\
    F11   & 3.96E+01 (1.05E+00) & 100.0\% (0.0\%) & \textbf{2.34E+01 (1.61E+00) +} & 100.0\% (0.0\%) = & 7.31E+01 (1.44E+00) & 100.0\% (0.0\%) & \textbf{4.42E+01 (2.75E+00) +} & 100.0\% (0.0\%) = \\
    F12   & 8.14E+05 (7.26E+04) & 100.0\% (0.0\%) & \textbf{3.44E+05 (4.11E+04) +} & 100.0\% (0.0\%) = & 4.08E+06 (2.88E+05) & 100.0\% (0.0\%) & \textbf{1.59E+06 (1.56E+05) +} & 100.0\% (0.0\%) = \\
    F13   & 4.33E+05 (3.79E+05) & 100.0\% (0.0\%) & \textbf{9.18E+04 (2.98E+04) +} & 100.0\% (0.0\%) = & \textbf{4.64E+05 (1.57E+05)} & 100.0\% (0.0\%) & 1.02E+06 (2.22E+05) - & 100.0\% (0.0\%) = \\
    F14   & \textbf{1.36E+01 (1.20E-01)} & \textbf{100.0\% (0.0\%)} & 1.38E+01 (1.30E-01) - & 80.4\% (0.6\%) - & \textbf{2.33E+01 (1.98E-01)} & \textbf{100.0\% (0.0\%)} & 2.36E+01 (1.33E-01) - & 91.8\% (0.3\%) - \\
    F15   & \textbf{6.15E+02 (4.79E+01)} & 100.0\% (0.0\%) & 6.53E+02 (4.75E+01) - & 100.0\% (0.0\%) = & 7.55E+02 (8.09E+01) & 100.0\% (0.0\%) & 7.87E+02 (1.53E+01) = & 100.0\% (0.0\%) = \\
    F16   & \textbf{3.15E+02 (2.82E+01)} & 100.0\% (0.0\%) & 3.39E+02 (2.99E+01) - & 100.0\% (0.0\%) = & \textbf{4.10E+02 (2.07E+01)} & 100.0\% (0.0\%) & 4.30E+02 (1.36E+01) - & 100.0\% (0.0\%) = \\
    F17   & \textbf{3.54E+02 (1.73E+01)} & 100.0\% (0.0\%) & 3.73E+02 (4.15E+01) - & 100.0\% (0.0\%) = & \textbf{4.80E+02 (2.92E+01)} & 100.0\% (0.0\%) & 4.90E+02 (2.69E+01) - & 100.0\% (0.0\%) = \\
    F18   & 9.45E+02 (4.99E+00) & 100.0\% (0.0\%) & 9.45E+02 (4.78E+00) = & 100.0\% (0.0\%) = & \textbf{1.05E+03 (1.19E+01)} & 100.0\% (0.0\%) & 1.08E+03 (1.41E+01) - & 100.0\% (0.0\%) = \\
    F19   & \textbf{9.43E+02 (4.62E+00)} & 100.0\% (0.0\%) & 9.47E+02 (5.39E+00) - & 100.0\% (0.0\%) = & \textbf{1.05E+03 (1.03E+01)} & 100.0\% (0.0\%) & 1.08E+03 (1.35E+01) - & 100.0\% (0.0\%) = \\
    F20   & 9.44E+02 (4.15E+00) & 100.0\% (0.0\%) & 9.45E+02 (6.06E+00) = & 100.0\% (0.0\%) = & \textbf{1.04E+03 (1.03E+01)} & 100.0\% (0.0\%) & 1.07E+03 (1.44E+01) - & 100.0\% (0.0\%) = \\
    F21   & 1.13E+03 (1.01E+01) & 100.0\% (0.0\%) & 1.13E+03 (6.15E+00) = & 100.0\% (0.0\%) = & 1.10E+03 (7.14E+00) & 100.0\% (0.0\%) & \textbf{1.08E+03 (6.40E+00) +} & 100.0\% (0.0\%) = \\
    F22   & \textbf{1.00E+03 (2.29E+01)} & 100.0\% (0.0\%) & 1.07E+03 (3.51E+01) - & 100.0\% (0.0\%) = & \textbf{1.03E+03 (1.27E+01)} & 100.0\% (0.0\%) & 1.06E+03 (1.83E+01) - & 100.0\% (0.0\%) = \\
    F23   & 1.13E+03 (1.14E+01) & 100.0\% (0.0\%) & 1.13E+03 (1.13E+01) = & 100.0\% (0.0\%) = & 1.10E+03 (5.51E+00) & 100.0\% (0.0\%) & \textbf{1.08E+03 (5.31E+00) +} & 100.0\% (0.0\%) = \\
    F24   & 1.03E+03 (8.14E+00) & 100.0\% (0.0\%) & 1.03E+03 (9.44E+00) = & 100.0\% (0.0\%) = & \textbf{1.12E+03 (1.23E+01)} & 100.0\% (0.0\%) & 1.13E+03 (1.33E+01) - & 100.0\% (0.0\%) = \\
    F25   & 1.96E+03 (2.52E+01) & 0.0\% (0.0\%) & \textbf{1.91E+03 (2.51E+01) +} & 0.0\% (0.0\%) = & 2.09E+03 (3.27E+01) & 0.0\% (0.0\%) & \textbf{2.03E+03 (3.78E+01) +} & 0.0\% (0.0\%) = \\
    \midrule
    +/=/- &       &       & 5/8/12 & 0/17/8 &       &       & 7/3/15 & 0/17/8 \\
    \bottomrule
    \end{tabular}%
  \label{tab:result2}%
\\The symbols "+/=/-" indicate that CVT-ME performed significantly better ($+$), not significantly better or worse ($=$), or significantly worse ($-$) compared to CVT-DME using the Wilcoxon rank sum test with $\alpha = 0.05$ significance level.
\end{table*}%

\section{Discussion and conclusions}

In this paper, we described Differential-MAP-Elites, a new algorithm combining elements of Differential Evolution with CVT-MAP-Elites, which is a version of MAP-Elites. The core idea of the algorithm is simple: the operators are taken from Differential Evolution, everything else from CVT-MAP-Elites.

Our goal in designing this algorithm was to combine the real-valued optimization performance of Differential Evolution with the behavior space exploration capability of CVT-MAP-Elites, and so creating a quality-diversity algorithm that would achieve higher-fit solutions and explore more of behavior space. The results suggest we succeeded. We tested Differential MAP-Elites versus CVT-MAP-Elites on a number of benchmarks from the classic CEC 2005 test set, for which we had created behavior characteristics from linear projections, and the former algorithm, on the whole, achieved higher-fit elites and much more coverage.

The results also suggest further investigation of this algorithm, including testing its performance on important problems with continuous representations such as training neural networks. There are also many existing modifications and improvements to the basic Differential Evolution algorithm, which should be explored in the context of Differential MAP-Elites.

Presumably, other QD algorithms could relatively easily be combined with Differential Evolution in order to improve real-valued optimization performance. In particular, the standard (grid-based) MAP-Elites could easily be combined with Differential Evolution. Possibly this would work well with e.g., Novelty Search with Local Competition as well.

\begin{acks}
This work was supported by the National Research Foundation of Korea (NRF) grant funded by the Korea government (MSIT) (No. 2020R1A2C1103138) and the National Science Foundation (NSF) award (No. 1717324).
\end{acks}

\bibliographystyle{ACM-Reference-Format}
\bibliography{references} 

\end{document}